\documentclass[letterpaper]{article} 
\usepackage{aaai25}  
\usepackage{times}  
\usepackage{helvet}  
\usepackage{courier}  
\usepackage[hyphens]{url}  
\usepackage{graphicx} 
\usepackage{natbib}  
\usepackage{caption} 
\usepackage{algorithm}
\usepackage{algorithmic}
\usepackage{booktabs}
\usepackage{amsmath}

\usepackage{newfloat}
\usepackage{enumitem}
\usepackage{listings}
\DeclareCaptionStyle{ruled}{labelfont=normalfont,labelsep=colon,strut=off} 
\floatstyle{ruled}
\newfloat{listing}{tb}{lst}{}
\floatname{listing}{Listing}
\title{LLM Rationalis? Measuring Bargaining Capabilities of AI Negotiators}

\author {
    Cheril Shah\textsuperscript{\rm 1},
    Akshit Agarwal\textsuperscript{\rm 1},
    Kanak Garg\textsuperscript{\rm 2},
    Mourad Heddaya\textsuperscript{\rm 3}
}
\affiliations {
    \textsuperscript{\rm 1}University of California San Diego\\
    \textsuperscript{\rm 2}University of Washington\\
    \textsuperscript{\rm 3}University of Chicago\\
    Correspondence to c4shah@ucsd.edu, aka002@ucsd.edu
}

\begin{document}

\maketitle

\begin{abstract}
Bilateral negotiation is a complex, context‐sensitive task in which human negotiators dynamically adjust anchors, pacing, and flexibility to exploit power asymmetries and informal cues. We introduce a unified mathematical framework for modeling concession dynamics based on a hyperbolic tangent curve, and propose two metrics burstiness ($\tau$) and the Concession‑Rigidity Index (CRI) to quantify the timing and rigidity of offer trajectories. We conduct a large‑scale empirical comparison between human negotiators and four state‑of‑the‑art large language models (LLMs) across natural‑language and numeric‑offers settings, with and without rich market context, as well as six controlled power‑asymmetry scenarios. Our results reveal that, unlike humans who smoothly adapt to situations and infer the opponents position and strategies, LLMs systematically anchor at extremes of the possible agreement zone for negotiations and optimize for fixed points irrespective of leverage or context. Qualitative analysis further shows limited strategy diversity and occasional deceptive tactics used by LLMs. Moreover the ability of LLMs to negotiate does not improve with better models. These findings highlight fundamental limitations in current LLM negotiation capabilities and point to the need for models that better internalize opponent reasoning and context‑dependent strategy.
\end{abstract}

%

\section{Introduction}

Bilateral bargaining scenarios involve a dynamic interplay of reasoning and communication, as each participant works to understand the other’s intentions and perspectives. Such understanding is essential for crafting strategic offers and employing persuasive language to steer negotiations toward mutually beneficial outcomes.

There is growing interest in leveraging large language models (LLMs) for negotiation tasks, both to support human training and to autonomously conduct economic interactions. Studying the negotiation capabilities of LLMs not only aids in deploying them in practical settings but also serves as a valuable lens to evaluate their underlying competencies. These include their ability to reason about incentives and goals, sustain coherent multi-turn dialogue, follow strategic prompts, and adapt to various roles and objectives.

In this work, we contribute to this emerging area by:
\begin{enumerate}[itemsep=0pt, parsep=0pt, topsep=0pt, partopsep=0pt]
  \item Proposing a mathematical framework and novel metrics to track offer dynamics and latent trends in negotiation settings;
  \item Comparing human and LLM negotiation performance under identical conditions;
  \item Investigating the role of context in shaping LLM negotiation behavior;
  \item Introducing controlled power asymmetries to assess their effects on outcomes;
  \item Conducting qualitative analysis of emergent strategies and linguistic patterns.
\end{enumerate}

\section{Related Work}

Recent research has begun to explore the economic behavior and strategic reasoning capabilities of LLMs in negotiation contexts.

\citep{ross2024llm} examined whether LLMs exhibit human-like behavioral biases by adapting canonical games from behavioral economics. They quantified biases such as inequity aversion, risk/loss aversion, and time discounting, and found that LLMs exhibit distinct behavioral patterns showing stronger altruism but weaker loss aversion compared to both humans and rational agents. However, their work involved fitting different utility curves to different games, without a general negotiation framework.

Another line of research has analyzed negotiation outcomes and tactics employed by LLMs. \citep{vaccaro-etal-2025-competition} showed that LLM agents perceived as “warm” reached agreements more frequently and were better at value creation in integrative negotiations. \citep{bianchi-etal-2024-negotiationarena} demonstrated that behavioral cues can improve agent payoffs by up to 20\%, while also revealing irrational tendencies. \citep{xia-etal-2024-measuring} highlighted the difficulties LLMs encounter when acting as buyers.

Despite these advances, existing studies rarely compare LLMs with humans in matched scenarios or explore how LLM behavior shifts across different negotiation structures, such as power asymmetry. They also tend to overlook qualitative aspects like language use and strategy emergence, and typically lack systematic quantitative tools to measure trends like concession patterns or inferred intentions.

Our work addresses these gaps by combining rigorous experimental control with both qualitative and quantitative analyses of negotiation dialogues.

\section{Negotiation setting}
\label{negset}

We adopted a bilateral negotiation scenario from \citep{heddaya-etal-2023-language}. In this setup, both the buyer and seller were informed of the \$240,000 asking price and shared identical information regarding the house, its surrounding area, and recent sales prices of comparable homes. Crucially, each participant also received a private valuation for the house: \$235,000 for the buyer and \$225,000 for the seller. 

To examine the role of information exchange, we defined two settings: (i) a numeric-only format, where parties exchanged numerical bids; and (ii) a natural language format, where negotiation occurred through free-form text, following \citep{heddaya-etal-2023-language}.

Human negotiation data for this setting was sourced from the dataset provided by \citep{heddaya-etal-2023-language}. For LLM negotiation data, we use similar prompts as given to humans to simulate $100$ self-play negotiations (negotiations where buyer and seller agent are simulated by the same LLM model) per model using the models GPT-o4-mini, GPT-4.1-mini, GPT-4o-mini and GPT-4.1-nano (\cite{openai2025o3o4-mini}, \cite{openai2025GPT4.1}, \cite{openai2024GPT4o-mini}).

\section{Modeling Negotiations}

Classical alternating--offers work (e.g.\ \citep{faratin1998negotiation}) model concessions with a power--law
\[
  p(t)=p_{\min}+(p_{\max}-p_{\min})\,t^{1/e},
\]
where the exponent $e$ yields linear $(e{=}1)$, early--concession (``conceder'', $e{<}1$), or late--concession (``boulware'', $e{>}1$) profiles.
As curvature is governed by a single parameter, the function cannot simultaneously capture richer patterns observed in bounded negotiations such as an early\-rigidity $\to$ mid\-stage\-flexibility $\to$ late\-rigidity arc, an early\-flexibility $\to$ late\-rigidity arc, or persistent rigidity throughout \citep{DBLP:conf/atal/BaarslagHHIJ14,oprea2002adaptive,RePEc:spr:grdene:v:15:y:2006:i:2:d:10.1007_s10726-006-9028-8}.

In addition, negotiators’ perceived reservation prices may diverge from the true $p_{\min},p_{\max}$ supplied ex--ante, further undermining the static power--law assumption.

To address this need for a model that can accommodate such nuanced dynamics, we therefore introduce a hyperbolic tangent model,

\[
  y(x)=d + b\,\tanh\bigl(a\,x - c\bigr),
\]
specifically focusing on its behavior in the first quadrant (representing non‑negative negotiation rounds $x$ and offer values $y$).
Critically, we fit separate tanh curves for buyers and sellers. This separate fitting allows the distinct parameters $(a,b,c,d)$ for each role to capture their unique strategies.

\begin{figure}
    \centering
    \includegraphics[width=1.0\linewidth]{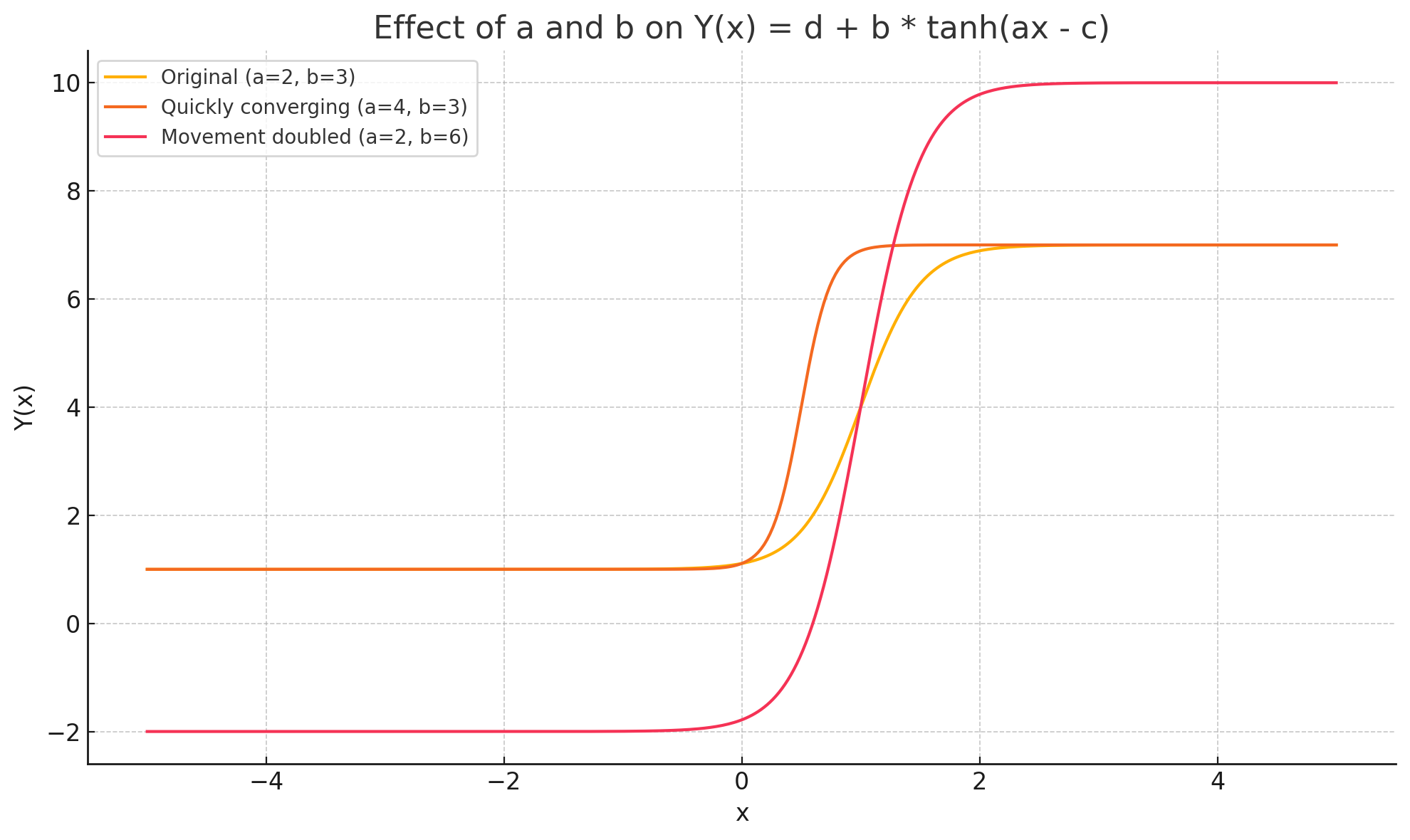}
    \caption{Effect of concession pace and concession span over negotiations}
    \label{fig:Concession_Pace_effects}
\end{figure}

Where,
\begin{itemize}
  \item \textbf{$a$: concession pace.}\
        Controls how quickly offers shift.  Larger $|a|$ compresses the high‑curvature region into a shorter interval (width $\approx1.32/|a|$).
        The sign indicates direction: $a>0$ implies upward movement of offers, $a<0$ downward.

  \item \textbf{$b$: concession span.}\
        Half of the total movement the negotiator is willing to make; the full range is $2|b|$. Figure~\ref{fig:Concession_Pace_effects} shows the effect of doubling $a$ versus doubling $b$ on the curve shape, ceteris paribus.

  \item \textbf{$d$: anchor point.}\
        Central target around which the negotiator’s offers oscillate.

  \item \textbf{$c$: horizontal shift.}\
        Controls the round index at which the curve’s steepest change is centered.
\end{itemize}

By computing the second derivative of $f(x)=d+b\tanh(ax-c)$ and setting it to zero, the elbow points—where the curve bends most sharply—satisfy
\[
  \mathrm{sech}^2(ax-c) = \tfrac{1}{2},
\]
which yields
\[
  x = \frac{c}{a} \pm \frac{\mathrm{arccosh}(\sqrt{2})}{a}\approx \frac{c}{a} \pm \frac{0.66}{a}.
\]
This ``elbow window'' defines the interval in which concessions occur at maximum speed.

The negotiation's \emph{burstiness} $\tau$ is defined as the peak concession rate,
\[
  \tau = |a_{\mathrm{scaled}}| \times b_{\mathrm{scaled}},
\]
where $a_{\mathrm{scaled}}$ and $b_{\mathrm{scaled}}$ are obtained via min--max normalization of the raw parameters across all fitted negotiations, ensuring each lies in $(0,1)$ and contributes equally.

To quantify the proportion of negotiation time spent in rapid concessions, we define the Concession--Rigidity Index (CRI)
\[
  \mathrm{CRI} = 1 - \frac{1.32}{|a|\,T},
\]
where $T$ is the total number of negotiation rounds.  Hence $\mathrm{CRI}\in[0,1]$, with values near 1 indicating a brief, intense burst of concessions (high rigidity) and values near 0 corresponding to steady concessions throughout (low rigidity).  As a single summary statistic, CRI captures the overall rigidity of the negotiation trajectory. We define a novel, data-driven Concession Rigidity Index (CRI\(^*\)) to quantify concession dynamics, and employ a multi-stage pipeline for clustering negotiation strategies. Complete methodology and validation details are provided in Appendix~\ref{sec:appendix}.

We fit a curve where $x$ the turn index and $y$ is the offer exchanged at $x$

The parameters $(\hat{a}, \hat{b}, \hat{c}, \hat{d})$ are estimated using non-linear least squares. This involves finding the parameter values that minimize the sum of the squared differences between the observed data points $y_i$ and the values predicted by the model function $f(x_i; a,b,c,d)$:

\begin{flalign*}
(\hat{a},\hat{b},\hat{c},\hat{d}) &= \arg\min_{a,b,c,d}
      \sum_i \bigl(y_i - f(x_i; a,b,c,d)\bigr)^2, &&\\[4pt]
f(x_i; a,b,c,d)                     &= d + b\,\tanh\!\bigl(a x_i - c\bigr). &&
\end{flalign*}

We fit the curves separately for each of the 100 negotiations per model and used the median of the fitted parameters to represent each model. More details about the curve fits can be found in the appendix \ref{sec:appendix}

\begin{figure}
    \centering
    \includegraphics[width=1\linewidth]{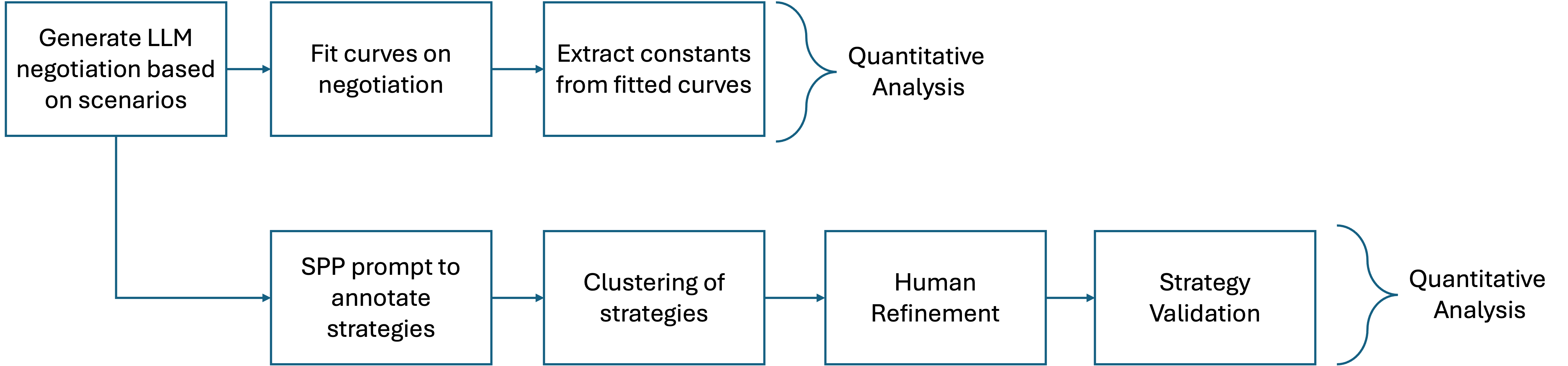}
    \caption{Multi-stage Pipeline}
    \label{fig:placeholder}
\end{figure}

\section{Results}
To understand how language models behave across negotiation settings, we evaluate performance under two interaction protocols: \textbf{(A) Natural Language} and \textbf{(B) Alternating Offers Only}.

\paragraph{Zone of Possible Agreement (ZOPA).}
Given a seller reservation price of \$225{,}000 (minimum acceptable) and a buyer reservation price of \$235{,}000 (maximum payable), the Zone of Possible Agreement spans \$225{,}000–\$235{,}000, with a midpoint at \$230{,}000. This range defines the theoretical bounds for settlement and guides anchor placement and concession dynamics.

For each configuration, we report the following key behavioral metrics:

\begin{itemize}
    \item \textbf{Anchoring distance ($d$):} The initial offer made by the buyer/seller relative to the center of the Zone of Possible Agreement (ZOPA). It reflects initial aggressiveness or conservatism in positioning.
    \item \textbf{Burstiness ($\tau$):} The degree to which offers change intermittently rather than gradually indicating strategic pacing of concessions.
    \item \textbf{Rigidity (CRI):} The fraction of turns in which an agent refuses to concede or repeats a previous offer, reflecting inflexibility.
    \item \textbf{Offer rounds ($T$):} The number of offer–counteroffer turns until deal or termination.
\end{itemize}

\subsection{Negotiating in Natural Language}

Table~\ref{tab:offer_params_original_nl} summarizes the results.

\begin{table*}[ht]
  \centering
  \caption{Natural Language Negotiation Results: ZOPA = \$225k–\$235k. Human metrics shown for comparison.}
  \label{tab:offer_params_original_nl}
  \resizebox{\textwidth}{!}{
  \begin{tabular}{lcccccccccc}
    \toprule
    \textbf{Agent} & \textbf{Role} &
    \textbf{Median Deal (\$k)} & \textbf{IQR} &
    \textbf{Anchor (\$k)} & \textbf{IQR\$k} &
    \textbf{Burstiness ($\tau$)} & \textbf{IQR} &
    \textbf{CRI} & \textbf{IQR} &
    \textbf{Turns ($T$)} \\
    \midrule
    Human         & Buyer  & 230.0 & 1.0 & 230.5 & 3.0 & 0.39 & 0.03 & 0.64 & 0.07 & 5.6 \\
                  & Seller & 230.0 & 1.7 & 229.5 & 2.3 & 0.51 & 0.09 & 0.72 & 0.02 & 5.6 \\
    \midrule
    GPT-4.1-mini  & Buyer  & 225.0 & 1.4 & 225.0 & 1.8 & 0.18 & 0.04 & 0.008 & 0.03 & 5.0 \\
                  & Seller & 225.0 & 1.7 & 228.0 & 4.5 & 0.47 & 0.05 & 0.58 & 0.06 & 5.0 \\
    GPT-4.1-nano  & Buyer  & 228.5 & 1.8 & 225.0 & 2.0 & 0.29 & 0.07 & 0.39 & 0.04 & 6.2 \\
                  & Seller & 228.5 & 1.7 & 227.5 & 1.9 & 0.49 & 0.06 & 0.61 & 0.05 & 6.2 \\
    GPT-4o-mini   & Buyer  & 225.0 & 1.9 & 225.0 & 2.1 & 0.27 & 0.05 & 0.22 & 0.07 & 5.3 \\
                  & Seller & 225.0 & 2.0 & 227.0 & 2.5 & 0.51 & 0.07 & 0.74 & 0.13 & 5.3 \\
    GPT-o4-mini   & Buyer  & 226.3 & 1.8 & 225.0 & 2.2 & 0.25 & 0.06 & 0.11 & 0.04 & 5.8 \\
                  & Seller & 226.3 & 1.9 & 226.5 & 1.8 & 0.50 & 0.11 & 0.56 & 0.05 & 5.8 \\
    \bottomrule
  \end{tabular}
  }
\end{table*}

\paragraph{Anchor Behavior.}
Human negotiators exhibit anchors near the midpoint of ZOPA (\$229.5–\$230.5), indicating mutual recognition of bargaining range. In contrast, all LLM buyers anchors uniformly at the seller’s floor (\$225k), reflecting a failure to assert value or infer strategic room. Sellers using GPT-4.1-nano, GPT-4.1-mini, and GPT-o4-mini often disclose reservation prices early, violating instructions and narrowing the effective ZOPA.

\paragraph{Concession Dynamics.}
Humans demonstrate sharp concession bursts and sustained rigidity, aligning with strategic patience (\(\tau\approx0.39{-}0.51\), CRI~$\approx$~0.64–0.72). GPT-4.1-mini's buyer exhibits the flattest concession curve and negligible rigidity (CRI = 0.008), suggesting over-compliance. GPT-4.1-nano performs more human-like timing but still lacks pacing control. Notably, GPT-4o-mini's seller is even more rigid than humans (CRI = 0.74), while GPT-o4-mini is most flexible (CRI = 0.56).

\paragraph{Negotiation Outcomes.}
Humans consistently settle at ZOPA midpoint (\$230k), balancing interests. GPT-4.1-mini and GPT-4o-mini gravitate to \$225k regardless of role, suggesting static target optimization. GPT-4.1-nano reaches higher settlements (\$228.5k), but still lacks bidirectional strategy. Overall, LLMs exhibit rigidity or over-compliance based on configuration, unable to shift anchors balance interests.

\paragraph{Qualitative Analysis.}
Anchoring \& Gradual Concession was the top strategy for GPT-4.1-nano (50\%), GPT-4o-mini (34\%), and humans (18\%). GPT-4.1-mini and GPT-o4-mini leaned on Rapport Building \& Expectation Management. Humans favored Active Listening \& Empathetic Probing (30\%), a strategy underused by all LLMs ($<5\%$). Interestingly, GPT-o4-mini fabricated BATNA 7\% of the time, followed by GPT-4.1-nano (5\%), humans (3\%), and others. GPT-o4-mini also engaged in Logrolling (6\%) as buyer.

\subsection{Negotiating with Alternating Offers Only}

Table~\ref{tab:offer_params_alternating} summarizes the results.

\begin{table*}[ht]
  \centering
  \caption{Alternating-Only Negotiation Results: ZOPA = \$225k–\$235k. Human metrics shown for comparison.}
  \label{tab:offer_params_alternating}
  \resizebox{\textwidth}{!}{
  \begin{tabular}{lcccccccccc}
    \toprule
    \textbf{Agent} & \textbf{Role} &
    \textbf{Median Deal (\$k)} & \textbf{IQR} &
    \textbf{Anchor (\$k)} & \textbf{IQR\$k} &
    \textbf{Burstiness ($\tau$)} & \textbf{IQR} &
    \textbf{CRI} & \textbf{IQR} &
    \textbf{Turns ($T$)} \\
    \midrule
    Human         & Buyer  & 230.0 & 0.9  & 229.8 & 1.5  & 0.36 & 0.04 & 0.60 & 0.04 & 5.1 \\
                  & Seller & 230.0 & 0.5  & 230.2 & 2.0  & 0.49 & 0.05 & 0.65 & 0.11 & 5.1 \\
    \midrule
    GPT-4.1-mini  & Buyer  & 225.0 & 1.6  & 225.0 & 1.7  & 0.19 & 0.04 & 0.04 & 0.03 & 4.8 \\
                  & Seller & 225.0 & 2.1  & 228.0 & 1.9  & 0.45 & 0.12 & 0.60 & 0.05 & 4.8 \\
    GPT-4.1-nano  & Buyer  & 228.0 & 1.8  & 225.0 & 2.2  & 0.31 & 0.09 & 0.33 & 0.04 & 5.6 \\
                  & Seller & 228.0 & 1.7  & 227.5 & 1.9  & 0.48 & 0.05 & 0.57 & 0.07 & 5.6 \\
    GPT-4o-mini   & Buyer  & 225.0 & 1.9  & 225.0 & 2.0  & 0.24 & 0.04 & 0.19 & 0.05 & 4.9 \\
                  & Seller & 225.0 & 2.1  & 227.0 & 2.3  & 0.50 & 0.07 & 0.71 & 0.06 & 4.9 \\
    GPT-o4-mini   & Buyer  & 226.0 & 1.8  & 225.0 & 1.7  & 0.27 & 0.05 & 0.12 & 0.04 & 5.2 \\
                  & Seller & 226.0 & 1.9  & 226.8 & 4.4 & 0.51 & 0.06 & 0.53 & 0.05 & 5.2 \\
    \bottomrule
  \end{tabular}
  }
\end{table*}

\paragraph{Anchor Behavior.}
Without natural language cues, all LLM buyers anchor rigidly at the floor (\$225k), showing no sign of inferred strategic value. Sellers for GPT-4.1-nano and GPT-o4-mini post slightly more assertive anchors (\$227.5k–\$228k), whereas GPT-4.1-mini and GPT-4o-mini again reveal their reservation prices early, reducing leverage. In contrast, human agents still nudge anchors near ZOPA midpoint, indicating implicit modeling of value even under numeric-only constraints.

\paragraph{Concession Dynamics.}
Human concession patterns retain their bursty nature (\(\tau\approx0.36{-}0.49\)) and moderate rigidity (CRI~$\approx$~0.60–0.65), aligning with competitive yet flexible pacing. GPT-4.1-mini exhibits minimal rigidity (CRI = 0.04) and shallow concession bursts (\(\tau = 0.19\)), suggesting a tendency to yield prematurely. GPT-4.1-nano displays a closer-to-human pacing profile but with lower rigidity, while GPT-4o-mini’s seller remains highly rigid (CRI = 0.71) across both protocols.

\paragraph{Negotiation Outcomes.}
Human agents reliably settle around the midpoint (\$230k), even without justification or persuasion tools. GPT-4.1-mini and GPT-4o-mini consistently close at the minimum (\$225k), while GPT-4.1-nano achieves better results (\$228k) but still fails to match human symmetry across roles. LLMs appear to either overfit to their own roles or lack bidirectional inference, leading to role-agnostic yet static settlements.

\subsection{Exploring Power Asymmetries in Negotiation}


To test the generalizability of our previous findings to different negotiation contexts, we systematically introduced power asymmetry into the negotiation scenarios. This was achieved by modifying the prompts to create six distinct negotiation scenarios, each characterized by different levels of assigned time pressure and Best Alternative To a Negotiated Agreement (BATNA) for the involved parties. Furthermore, we explored whether the provision of specific contextual information like details about the house under negotiation and comparable nearby properties affects the LLMs' reasoning processes. To this end, we created two distinct versions of the prompts: one incorporating this rich contextual information, and a second version that omitted these details, providing the LLM solely with its reservation price, BATNA, and time pressure constraints. Following our earlier experiment we also studied the behavior of LLMs when they negotiate using alternating offers only. 

\begin{table}[h!]
 \centering
 \caption{Power-Asymmetry Scenarios: (+1: Strong BATNA / Low time pressure; –1: Weak BATNA / High time pressure; 0: Neutral).}
 \label{tab:asymmetry_scenarios}
 \resizebox{0.9\columnwidth}{!}{
 \begin{tabular}{lcc}
 \toprule
 \textbf{Scenario} & \textbf{Seller Power} & \textbf{Buyer Power} \\
 \midrule
 1: Strong Seller & $+1$ / $-1$ & $+0$ / $+0$ \\
 2: Strong Buyer & $+0$ / $+0$ & $+1$ / $-1$ \\
 3: Weak Buyer & $+0$ / $+0$ & $-1$ / $+1$ \\
 4: Weak Seller & $-1$ / $+1$ & $+0 / +0$ \\
 5: Both Weak & $-1$ / $+1$ & $-1$ / $+1$ \\
 6: Both Strong & $+1$ / $-1$ & $+1$ / $-1$ \\
 \bottomrule
 \end{tabular}
 }
\end{table}

\begin{table*}[ht]
  \centering
  \caption{Natural Language Power-Asymmetry Outcomes (Median Deal Price in \$k). LLM results shown with/without context.}
  \label{tab:power_asymmetry_results}
  \resizebox{\textwidth}{!}{
  \begin{tabular}{lcccccc}
    \toprule
    \textbf{Scenario} & \textbf{GPT-4.1-mini (Ctx/NoCtx)} & \textbf{GPT-4.1-nano (Ctx/NoCtx)} & \textbf{GPT-4o-mini (Ctx/NoCtx)} & \textbf{GPT-o4-mini (Ctx/NoCtx)} & \textbf{Key Observation} \\
    \midrule
    1: Strong Seller & 232.5 / 225.0 & 235.0 / 232.5 & 225.0 / 225.0 & 226.3 / 225.0 & \textbf{LLMs ignore seller strength; anchors remain extreme.} \\
    2: Strong Buyer  & 230.0 / 225.0 & 235.0 / 231.0 & 225.0 / 227.5 & 235.0 / 225.0 & \textbf{LLMs respond inconsistently to buyer advantage.} \\
    3: Weak Buyer    & 233.8 / 225.0 & 235.0 / 235.0 & 225.0 / 225.0 & 233.8 / 225.0 & \textbf{LLMs show fixed bias; lack adaptation to weakness.} \\
    4: Weak Seller   & 232.5 / 225.0 & 235.0 / 232.5 & 225.0 / 225.0 & 235.0 / 225.0 & \textbf{LLMs ignore seller weakness; maintain high anchors.} \\
    5: Both Weak     & 233.8 / 225.0 & 235.0 / 231.0 & 225.0 / 225.0 & 225.0 / 225.0 & \textbf{LLMs default to extremes; no midpoint settlement.} \\
    6: Both Strong   & 230.0 / 225.0 & 233.8 / 235.0 & 225.0 / 225.0 & 225.0 / 225.0 & \textbf{LLMs collapse to extremes; fail to balance tension.} \\
    \bottomrule
  \end{tabular}
  }
\end{table*}


\subsubsection{Negotiation behaviour with context}
\label{subsubsec:Negotiation_behaviour_with_context}

\paragraph{Anchors ($d$).}
Across scenarios, LLMs ignored leverage: weak sellers still opened at extreme highs (\textit{e.g.}, GPT-4.1-series at \$235\,k) and strong buyers at extreme lows (\textit{e.g.}, GPT-4o-mini $<\$225\,k$).

\paragraph{Peak concession ($\tau$).}
LLMs conceded more when strong and less when weak. A salient case is GPT-o4-mini (buyer) with its highest $\tau\approx0.58$ in Scenario 2 despite holding the advantage.

\paragraph{Rigidity (CRI).}
Only GPT-4.1-nano showed leverage-sensitive rigidity (CRI $\uparrow$ to 0.71 when strong, $\downarrow$ to 0.54 when weak). Other models either stayed at similar rigidity levels or became even more rigid under weakness.

\paragraph{Final surplus split.}
Outcomes clustered at ZOPA edges: GPT-4.1-nano hit the seller-max \$235\,k in four of six scenarios, while GPT-4o-mini secured the buyer-max \$225\,k in three.

\paragraph{Qualitative Analysis} Most LLMs relied on the same negotiation tactics: anchoring, justification, and gradual concessions. This was especially true for GPT-4.1-nano, which showed little flexibility and repeated anchoring loops. In contrast, GPT-4.1-mini was the most versatile, mixing in rapport-building, strategic framing, and BATNA-awareness in multi-step negotiations. GPT-o4-mini leaned heavily on relational tactics, leading with rapport even when assertiveness might have been better possibly because of heavy post-training.

\subsubsection{Negotiation behaviour without contextual cues}
\label{subsubsec:Negotiation_behaviour_without_context}

\paragraph{Anchors ($d$).}
When blind to the scenario, most models fixate on the ZOPA edges: buyer agents of GPT-4.1-mini, GPT-o4-mini, and GPT-4o-mini repeatedly open at the \$225\,k floor, while their seller counterparts cluster near \$233\,k.  Only GPT-4.1-nano shows tempered anchors (\$228–230\,k as a buyer; \$232.5–233.8\,k as a seller), and GPT-4o-mini lowers its seller anchor to \$231.3\,k when (nominally) weak.

\paragraph{Peak concession ($\tau$).}
Leverage–sensitive timing largely disappears.  GPT-4.1-mini buyers stay rigid (\(\tau\!\approx\!0.2\) in every scenario), GPT-4.1-nano adapts (\(\tau\!\uparrow\!0.73\) when weak, \(\downarrow\!0.26\) when strong), whereas GPT-4o-mini swings from near-zero concessions as a seller (\(\tau\!=\!1.5\!\times\!10^{-4}\)) to surprisingly generous peaks as a buyer (\(\tau\!=\!0.78\) when already strong).

\paragraph{Rigidity (CRI).}
Patterns diverge: GPT-4.1-mini remains fully flexible (\(\mathrm{CRI}=0\)) despite low \(\tau\); GPT-4.1-nano adjusts its rigidity (0.74 when weak, 0.24 when strong); the other models oscillate between extremes—sellers of GPT-4.1-mini, GPT-4.1-nano, and GPT-o4-mini hover around 0.6, while GPT-4o-mini toggles from 0 to 0.68.

\paragraph{Final surplus split.}
Final surplus split outcomes polarize: GPT-4.1-mini, GPT-o4-mini, and GPT-4o-mini consistently converge toward the buyer-optimal price of \$225k across most runs. In contrast, GPT-4.1-nano reliably secures \$231–235k for sellers.

\paragraph{Qualitative Analysis.} Without context, LLMs struggled to adapt their negotiation strategies. Most buyer agents stuck to rigid anchoring at the seller’s minimum, especially GPT-4.1-nano. GPT-4.1-mini was somewhat more adaptable, using gradual concessions and rapport, while GPT-o4-mini defaulted to being cooperative even when the situation called for aggression. GPT-4o-mini had the least consistent approach, rarely using key assertive strategies like Anchoring \& BATNA Leverage or Strategic Framing (just 0.3–0.7\%). Deceptive tactics, like making up BATNAs, also showed up more when context was missing.

\subsubsection{Alternating Offers}

\paragraph{Peak concession ($\tau$).}
Buyers generally conceded little, with most models holding $\tau \approx 0.17\text{–}0.27$ across scenarios. GPT-4.1-nano showed slight leverage-based variation, GPT-4o-mini spiked only once (Scenario 3), and GPT-o4-mini occasionally made large concessions when advantaged or in specific contexts (peaking at $0.71$).  
Sellers showed more diversity: GPT-4.1-mini stayed moderate ($\tau\approx0.49\text{–}0.57$), GPT-4.1-nano was consistently low, GPT-4o-mini reluctant throughout, and GPT-o4-mini polarized near-zero when disadvantaged, moderate otherwise.

\paragraph{Rigidity (CRI).}
Buyer rigidity ranged from none (GPT-4.1-mini) to consistently high (GPT-4.1-nano). GPT-4o-mini spanned moderate to high, and GPT-o4-mini oscillated between no rigidity in some scenarios and high rigidity in others.  
Sellers similarly varied: GPT-4.1-mini stayed moderate, GPT-4.1-nano highly rigid, GPT-4o-mini mixed with occasional flexibility, and GPT-o4-mini swung between no rigidity and high rigidity depending on context.

\paragraph{Anchors ($d$).}
Buyers clustered at fixed points: GPT-4.1-mini at \$225\,k except for slight increases, GPT-4.1-nano consistently high (\$231–234\,k), GPT-4o-mini tightly around \$226.5\,k, and GPT-o4-mini at \$225\,k with modest bumps.  
Sellers split between midpoints (GPT-4.1-mini), consistently high anchors (GPT-4.1-nano, GPT-o4-mini), and lower, adaptive anchors (GPT-4o-mini).

\section{Conclusion}

This paper demonstrates that LLMs generally lack sophisticated, human-like negotiation strategies, tending to optimize a single aspect and producing overly buyer or seller-friendly outcomes regardless of scenario or context. We do this using a novel mathematical framework using a hyperbolic tangent model and metrics based off it. Their behavior changes somewhat with new information, but these changes are mostly specific to each model. For example, GPT-4.1-mini rarely adapts across scenarios, following a basic strategy unless both dialogue and context are present; then it sometimes makes large but poorly placed concessions. GPT-4.1-nano mainly responds to dialogue, sticking to a seller-favoring approach even without market facts, though it slightly softens its concessions. GPT-4o-mini is consistently buyer-oriented, but dialogue increases its concessions and removing facts adds volatility. Lastly, GPT-o4-mini is extremely sensitive to context. Furthermore, qualitative analysis revealed a significant gap in strategic diversity. While 30\% of human strategies involved 
Active Listening \& Empathetic Probing, this was underutilized by all LLMs ($<5\%$). More troublingly, some models engaged in deceptive tactics, such as fabricating BATNA claims, a behavior most prominent in GPT-o4-mini.

\bibliography{aaai25}

@inproceedings{
ross2024llm,
title={{LLM} economicus? Mapping the Behavioral Biases of {LLM}s via Utility Theory},
author={Jillian Ross and Yoon Kim and Andrew Lo},
booktitle={First Conference on Language Modeling},
year={2024},
url={https://openreview.net/forum?id=Rx3wC8sCTJ}
}

@article{bianchi-etal-2024-negotiationarena,
  title   = "{H}ow Well Can {LLM}s Negotiate? {NegotiationArena} Platform and Analysis",
  author  = "Bianchi, Federico and Chia, Patrick John and Yuksekgonul, Mert and Tagliabue, Jacopo and Jurafsky, Dan and Zou, James",
  journal = "arXiv preprint arXiv:2402.05863",
  year    = "2024"
}

@article{vaccaro-etal-2025-competition,
  title = "Advancing {AI} Negotiations: New Theory and Evidence from a Large-Scale Autonomous Negotiations Competition",
  author = "Vaccaro, Michelle and Caoson, Michael and Ju, Harang and Aral, Sinan and Curhan, Jared R.",
  journal = "arXiv preprint arXiv:2503.06416",
  year = "2025"
}

@misc{openai2025o3o4-mini,
  author       = {{OpenAI}},
  title        = {Introducing OpenAI o3 and o4-mini},
  howpublished = {\url{https://openai.com/index/introducing-o3-and-o4-mini/}},
  month        = apr,
  day          = {16},
  year         = {2025},
  note         = {Accessed: 2025-05-17}
}

@misc{openai2025gpt4.1,
  author       = {{OpenAI}},
  title        = {Introducing GPT-4.1 in the API},
  howpublished = {\url{https://openai.com/index/gpt-4-1/}},
  month        = apr,
  day          = {14},
  year         = {2025},
  note         = {Accessed: 2025-05-17}
}

@misc{openai2024gpt4o-mini,
  author       = {{OpenAI}},
  title        = {GPT-4o mini: advancing cost-efficient intelligence},
  howpublished = {\url{https://openai.com/index/gpt-4o-mini-advancing-cost-efficient-intelligence/}},
  month        = jul,
  day          = {18},
  year         = {2024},
  note         = {Accessed: 2025-05-17}
}

@inproceedings{DBLP:conf/atal/BaarslagHHIJ14,
  author={Tim Baarslag and Rafik Hadfi and Koen V. Hindriks and Takayuki Ito and Catholijn M. Jonker},
  title={Optimal Non-adaptive Concession Strategies with Incomplete Information},
  year={2014},
  cdate={1388534400000},
  pages={39-54},
  url={https://doi.org/10.1007/978-3-319-30307-9_3},
  booktitle={ANAC@AAMAS}
}

@article{oprea2002adaptive,
  author  = {Mihaela Oprea},
  title   = {An Adaptive Negotiation Model for Agent-Based Electronic Commerce},
  journal = {Studies in Informatics and Control},
  volume  = {11},
  number  = {3},
  pages   = {271--279},
  year    = {2002},
  url     = {https://sic.ici.ro/documents/995/Art._5_Issue_3_SIC_2002.pdf}
}

@article{faratin1998negotiation,
  author    = {Faratin, Peyman and Sierra, Carles and Jennings, Nicholas R.},
  title     = {Negotiation Decision Functions for Autonomous Agents},
  journal   = {Robotics and Autonomous Systems},
  year      = {1998},
  volume    = {24},
  number    = {3--4},
  pages     = {159--182},
  doi       = {10.1016/S0921-8890(98)00029-3},
  url       = {http://eprints.soton.ac.uk/252117/}
}

@article{wang2024unleashing,
  author    = {Wang, Zhenhailong and Mao, Shaoguang and Wu, Wenshan and Ge, Tao and Wei, Furu and Ji, Heng},
  title     = {Unleashing the Emergent Cognitive Synergy in Large Language Models: A Task-Solving Agent through Multi-Persona Self-Collaboration},
  journal   = {arXiv preprint arXiv:2307.05300},
  year      = {2024},
  url       = {https://arxiv.org/abs/2307.05300}
}

@Article{RePEc:spr:grdene:v:15:y:2006:i:2:d:10.1007_s10726-006-9028-8,
  author={Vivi Nastase},
  title={{Concession Curve Analysis for Inspire Negotiations}},
  journal={Group Decision and Negotiation},
  year=2006,
  volume={15},
  number={2},
  pages={185-193},
  month={March},
  doi={10.1007/s10726-006-9028-8},
  url={https://ideas.repec.org/a/spr/grdene/v15y2006i2d10.1007_s10726-006-9028-8.html}
}

@misc{Meng2024SFR-Embedding,
  title        = {SFR-Embedding-Mistral: Enhance Text Retrieval with Transfer Learning},
  author       = {Meng, Rui and Liu, Ye and Joty, Shafiq and Xiong, Caiming and Zhou, Yingbo and Yavuz, Semih},
  year         = {2024},
  month        = {Oct},
  day          = {28},
  howpublished = {\url{https://www.salesforce.com/blog/sfr-embedding/}},
  note         = {Salesforce AI Blog}
}

@inproceedings{xia-etal-2024-measuring,
    title = "Measuring Bargaining Abilities of {LLM}s: A Benchmark and A Buyer-Enhancement Method",
    author = "Xia, Tian  and
      He, Zhiwei  and
      Ren, Tong  and
      Miao, Yibo  and
      Zhang, Zhuosheng  and
      Yang, Yang  and
      Wang, Rui",
    editor = "Ku, Lun-Wei  and
      Martins, Andre  and
      Srikumar, Vivek",
    booktitle = "Findings of the Association for Computational Linguistics: ACL 2024",
    month = aug,
    year = "2024",
    address = "Bangkok, Thailand",
    publisher = "Association for Computational Linguistics",
    url = "https://aclanthology.org/2024.findings-acl.213/",
    doi = "10.18653/v1/2024.findings-acl.213",
    pages = "3579--3602"
}

@inproceedings{heddaya-etal-2023-language,
    title = "Language of Bargaining",
    author = "Heddaya, Mourad  and
      Dworkin, Solomon  and
      Tan, Chenhao  and
      Voigt, Rob  and
      Zentefis, Alexander",
    editor = "Rogers, Anna  and
      Boyd-Graber, Jordan  and
      Okazaki, Naoaki",
    booktitle = "Proceedings of the 61st Annual Meeting of the Association for Computational Linguistics (Volume 1: Long Papers)",
    month = jul,
    year = "2023",
    address = "Toronto, Canada",
    publisher = "Association for Computational Linguistics",
    url = "https://aclanthology.org/2023.acl-long.735/",
    doi = "10.18653/v1/2023.acl-long.735",
    pages = "13161--13185"
}

\vspace{2em}
\hrule
\vspace{2em}

\appendix
\section*{Appendix}
\label{sec:appendix}

\subsection*{Why a hyperbolic tangent?}  
While the traditional power–law form 
\[
  p(t) = p_{\min} + (p_{\max}-p_{\min})\,t^{1/e}
\]
provides a single “early vs.\ late concession” parameter via the exponent
\(e\), it suffers from two key limitations: (1) it can only produce monotonically decelerating or accelerating curves (no change in curvature sign), and (2) it ties the overall concession span and the curvature into one parameter.  In contrast, the hyperbolic tangent  
\[
  y(x)=d + b\,\tanh(a\,x - c)
\]
provides:

\begin{itemize}[noitemsep,topsep=0pt]
  \item \emph{Bounded asymptotes.}  Two finite limits \(d\pm b\), matching negotiators’ reservation prices and ensuring no “overshoot.”  
  \item \emph{Curvature control.} The shape naturally transitions from concave to convex and back (an “S–curve”), capturing early rigidity, mid-round flexibility, and late-stage rigidity within a single function.  
  \item \emph{Decoupled span vs.\ pace.}  Parameter \(b\) determines the total concession range, while \(a\) ontrols the steepness and timing—allowing adjustment of intensity independent of magnitude.
  
  \item \emph{Analytic tractability.}  Closed-form derivatives yield explicit “elbow” points and max-speed windows, enabling principled summary metrics (burstiness, CRI\(^*\)) without numerical approximations.  
\end{itemize}

These features make the tanh model both more expressive (able to recreate the richer concession profiles observed in practice) and more interpretable (clear, independent semantic roles for each parameter) than the single-exponent power–law.  

\subsection*{A Data-Driven Concession Rigidity Index (CRI\textsuperscript{*})}

Rather than relying on a single empirical constant, we measure \emph{rigidity} as the fraction of negotiation time spent in rapid concession.  Concretely:

\paragraph{1. Fit the negotiation curve}
\[
  y(x) = d + b\,\tanh\bigl(a\,x - c\bigr)
  \quad\text{to observed offers }(x_i,y_i).
\]

\paragraph{2. Instantaneous concession speed}
\[
  s(x) = |y'(x)| = |a\,b|\,\bigl[1 - \tanh^2(a\,x - c)\bigr].
\]

\paragraph{3. Normalized speed profile}
\[
  \hat s(x) = \frac{s(x)}{\displaystyle \max_{0\le x\le T} s(x)} \in [0,1].
\]

\paragraph{4. High--activity window}
Choose a threshold $\theta\in(0,1)$ (e.g. $\theta=0.1$).  Define
\[
  W = \{x\in[0,T]: \hat s(x)\ge\theta\},
  \quad
  \ell_W = \mathrm{length}(W).
\]
This $\ell_W$ is the total "active concession" time.

\paragraph{5. New CRI}
\[
  \mathrm{CRI}^* = 1 - \frac{\ell_W}{T}.
\]
- If $\ell_W\ll T$, then $\mathrm{CRI}^*\approx1$ (high rigidity).  
- If $\ell_W\approx T$, then $\mathrm{CRI}^*\approx0$ (low rigidity).

\paragraph{6. Proof that $0\le\mathrm{CRI}^*\le1$}
\[
  0 \le \ell_W \le T
  \quad\Longrightarrow\quad
  0 \le \frac{\ell_W}{T} \le 1
  \quad\Longrightarrow\quad
  0 \le 1 - \frac{\ell_W}{T} \le 1.
\]

\paragraph{7. Extremal behavior}
\begin{itemize}
  \item If $\hat s(x)\ge\theta$ for all $x$, then $\ell_W=T$ and $\mathrm{CRI}^*=0$.  
  \item If $\ell_W\to0$, then $\mathrm{CRI}^*\to1$.
\end{itemize}

\paragraph{8. Sensitivity to $b$}  
Since $s(x)\propto|a\,b|$, larger $|b|$ widens the region $\{\hat s\ge\theta\}$ and thus lowers $\mathrm{CRI}^*$.  Thus, $\mathrm{CRI}^*$ captures the combined effects of concession pace $(a)$ and span $(b)$.

\subsection*{Qualitative Analysis}
We employed a systematic multi-stage pipeline (Figure~\ref{fig:placeholder}) to extract and organize negotiation strategies across our datasets:
\begin{itemize}
\item \textbf{LLM Annotation:} We used GPT-4.1 with Solo Performance Prompting (SPP) \citep{wang2024unleashing} to label strategies at each turn by prompting four expert personas (Economist, Statistician, Linguist, Cognitive Scientist) to collaboratively refine strategic interpretations. Human evaluators rated SPP annotations as more accurate and diverse than standard or chain-of-thought prompting.

\item \textbf{Clustering:} We encoded 72 unique strategy labels using SFR-Embedding-Mistral \citep{Meng2024SFR-Embedding}, and grouped semantically similar strategies using hierarchical clustering (cosine similarity, complete linkage), selecting 22 clusters based on DBI minimization (which favors compact, well-separated clusters) and silhouette analysis (which quantifies intra- vs. inter-cluster cohesion).

\item \textbf{Human Refinement:} Annotators merged redundant clusters, corrected misclassifications, and filtered noise, resulting in 12 final strategy categories.

\item \textbf{Validation:} Three annotators independently reviewed a stratified sample, achieving Fleiss's $\kappa = 0.67$ (moderate inter-annotator agreement) and an average LLM-human agreement of 0.62, indicating consistent labeling quality.
\end{itemize}

\subsubsection*{Visualization: Negotiation Behaviour with Context}

The following plots illustrate negotiation dynamics across the six asymmetric power scenarios with context as discussed in Section~\ref{subsubsec:Negotiation_behaviour_with_context}.

\begin{figure}[h]
    \centering
    \includegraphics[width=\linewidth]{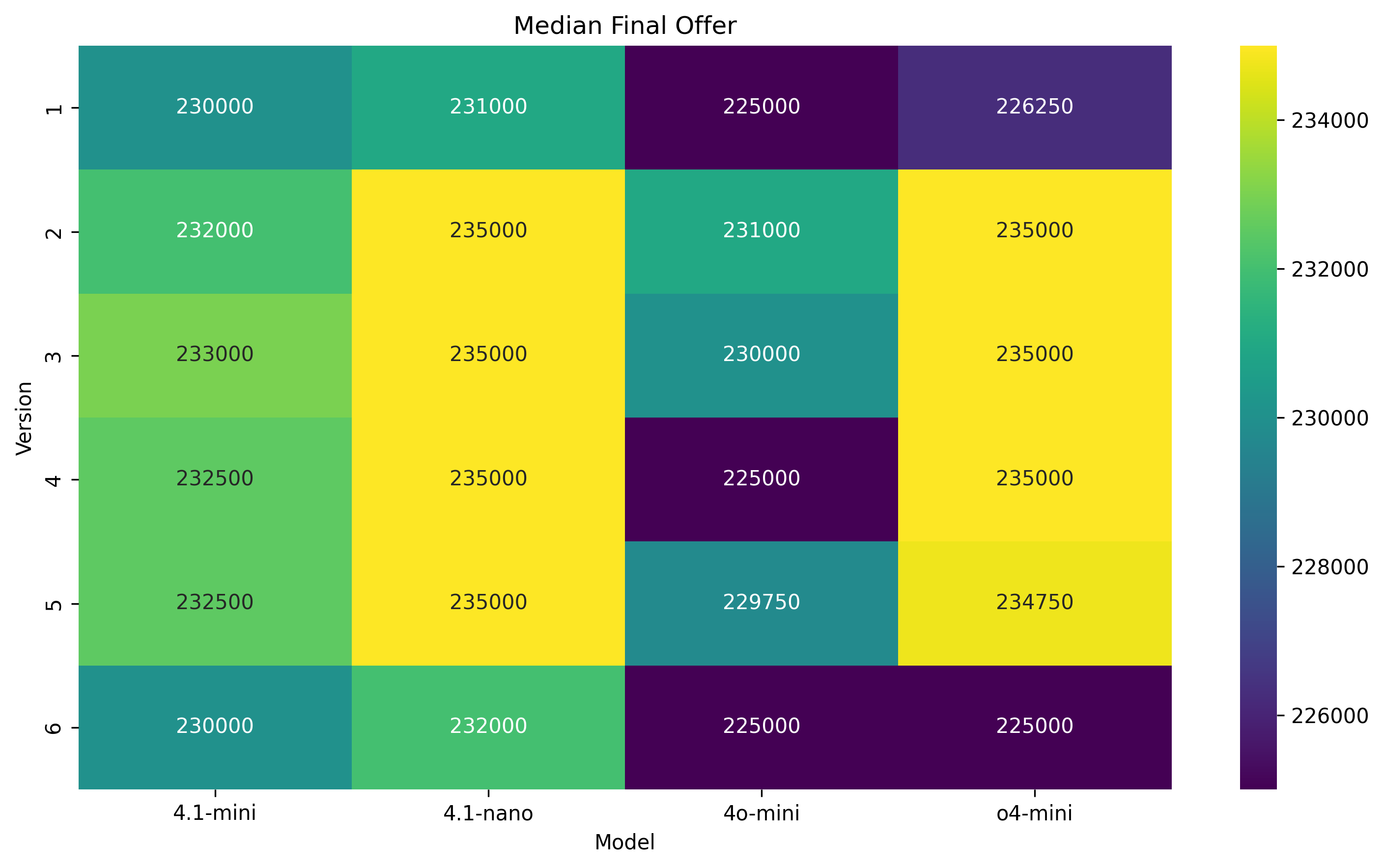}
    \caption{Final deal value (in \$) across power asymmetry scenarios. GPT-4.1-nano often secures seller-max outcomes (\$235k), while GPT-4o-mini typically converges at buyer-min values (\$225k).}
    \label{fig:appendix_median_final_offer}
\end{figure}

\begin{figure}[h]
    \centering
    \includegraphics[width=\linewidth]{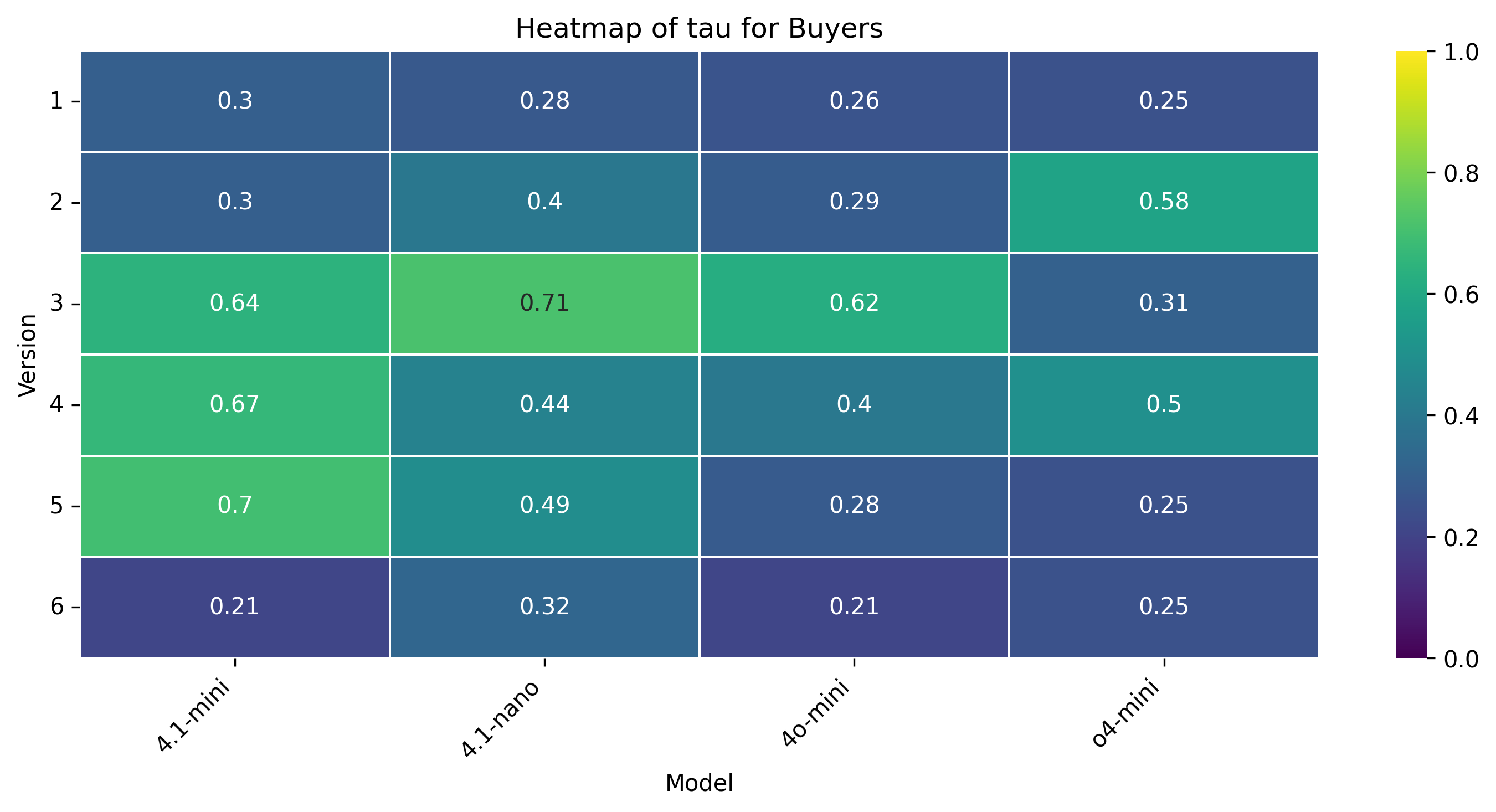}
    \caption{Buyer-side peak concession rate $\tau$. GPT-o4-mini exhibits its highest $\tau$ even when holding a strong position (Scenario 2), indicating poor leverage sensitivity.}
    \label{fig:appendix_tau_buyer}
\end{figure}

\begin{figure}[h]
    \centering
    \includegraphics[width=\linewidth]{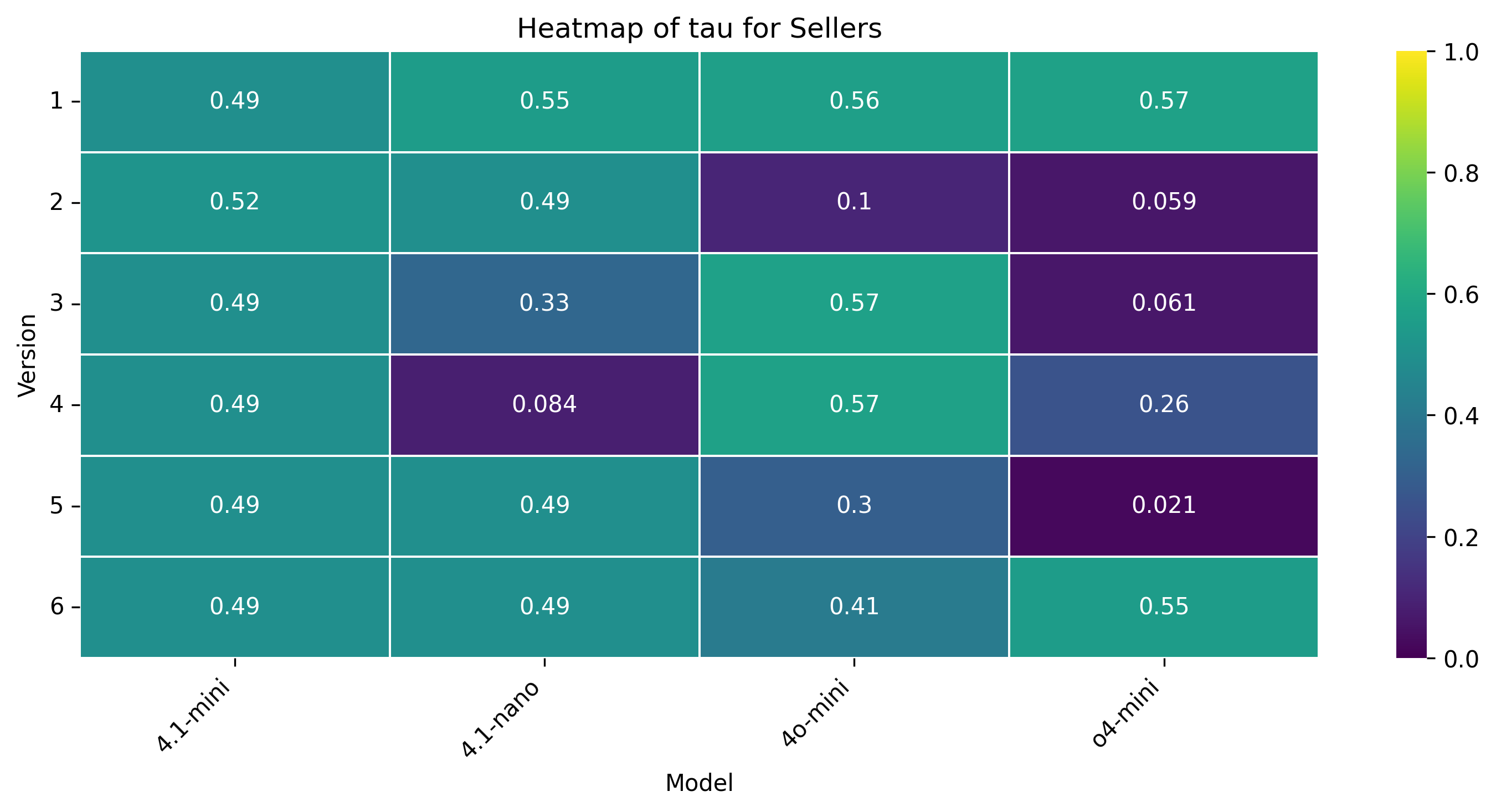}
    \caption{Seller-side peak concession rate $\tau$. Some models show rigidity even when weak; GPT-4.1-nano adapts slightly but inconsistently.}
    \label{fig:appendix_tau_seller}
\end{figure}

\begin{figure}[h]
    \centering
    \includegraphics[width=\linewidth]{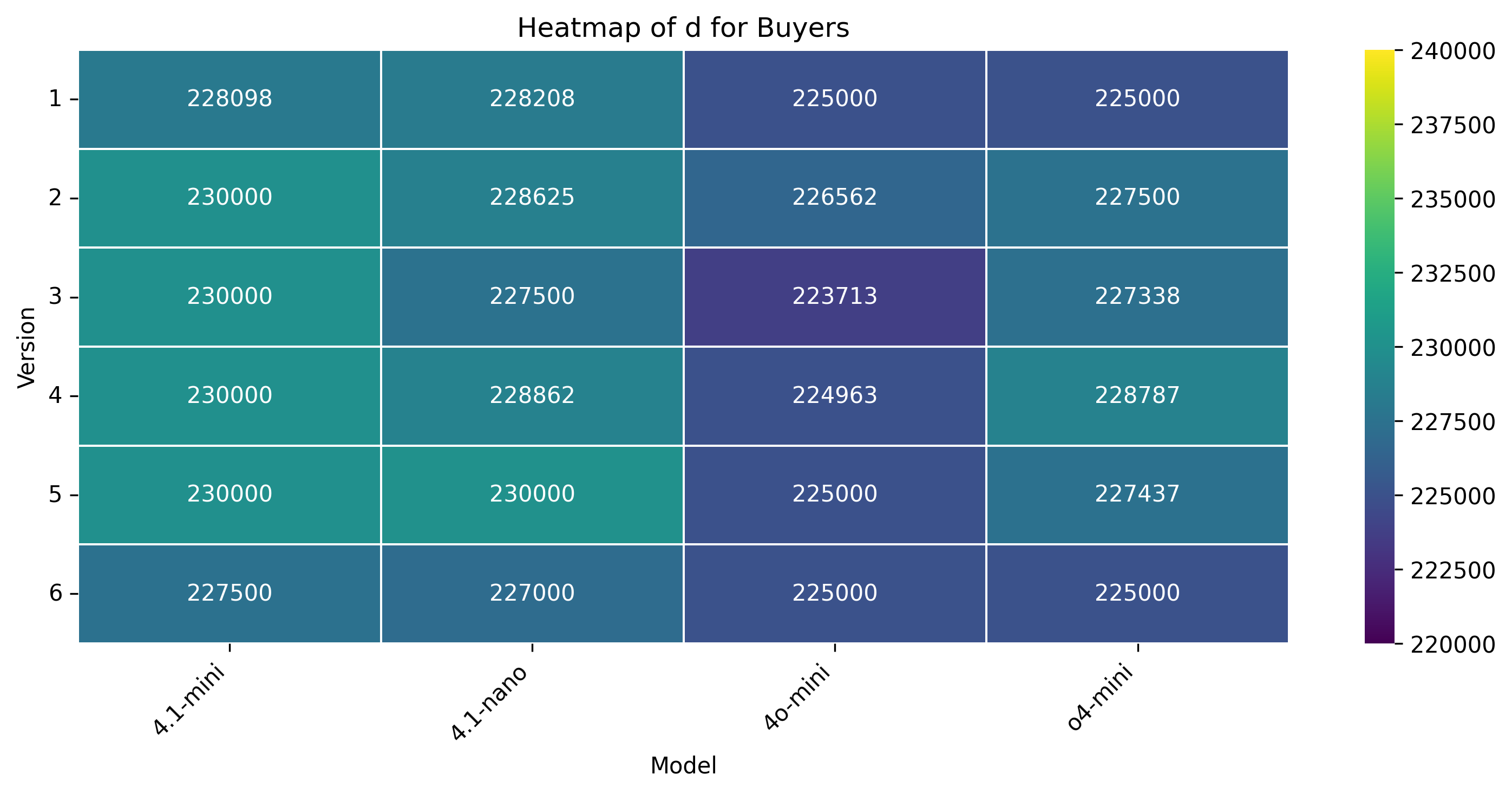}
    \caption{Buyer anchor values ($d$) across scenarios. Most models, including GPT-4.1-mini and GPT-o4-mini, anchor rigidly at the \$225k floor regardless of power position.}
    \label{fig:appendix_anchor_buyer}
\end{figure}

\begin{figure}[h]
    \centering
    \includegraphics[width=\linewidth]{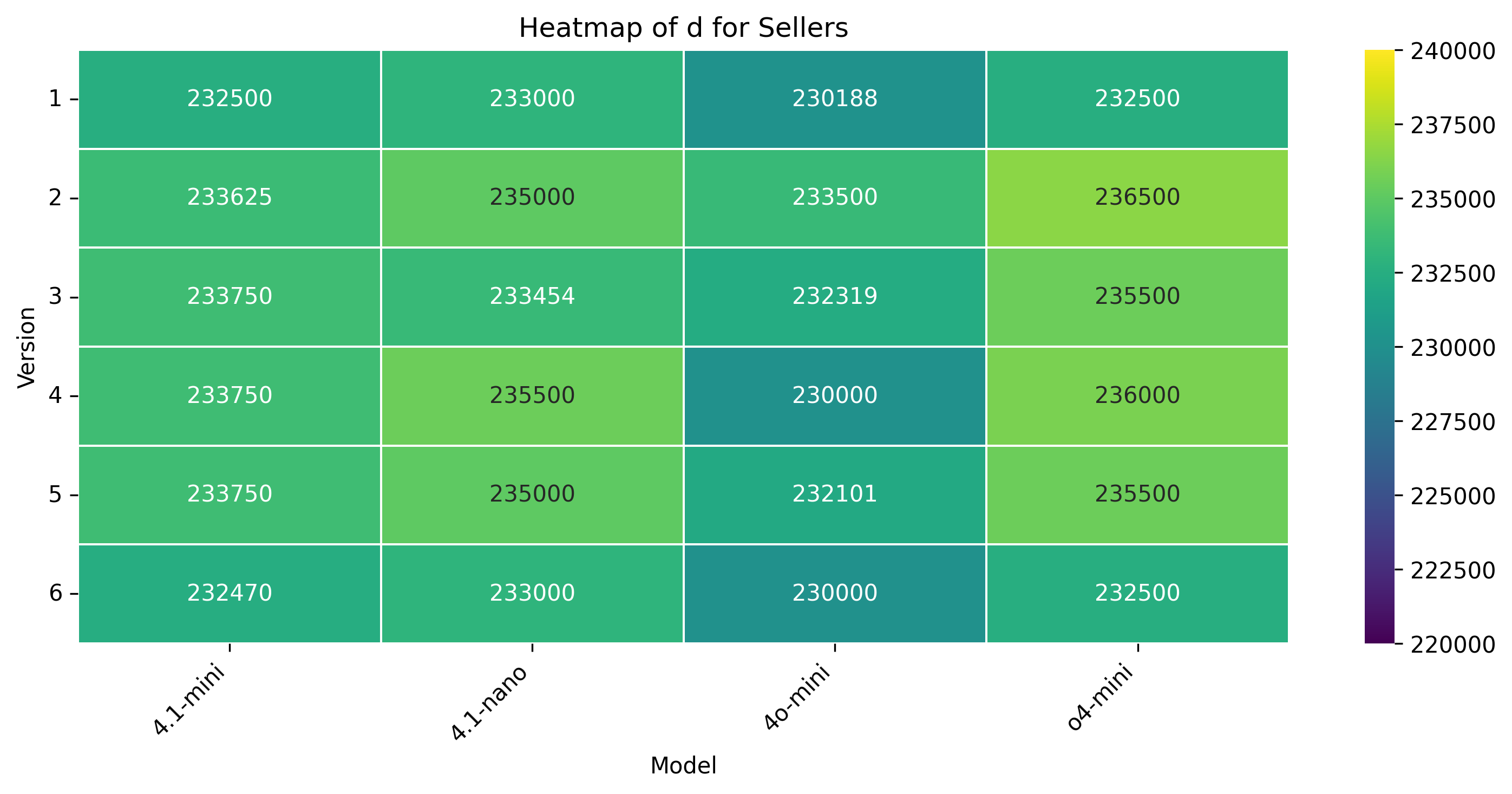}
    \caption{Seller anchor values ($d$). Some models show static behavior; GPT-4.1-nano consistently opens near the seller-max regardless of leverage.}
    \label{fig:appendix_anchor_seller}
\end{figure}

\begin{figure}[h]
    \centering
    \includegraphics[width=\linewidth]{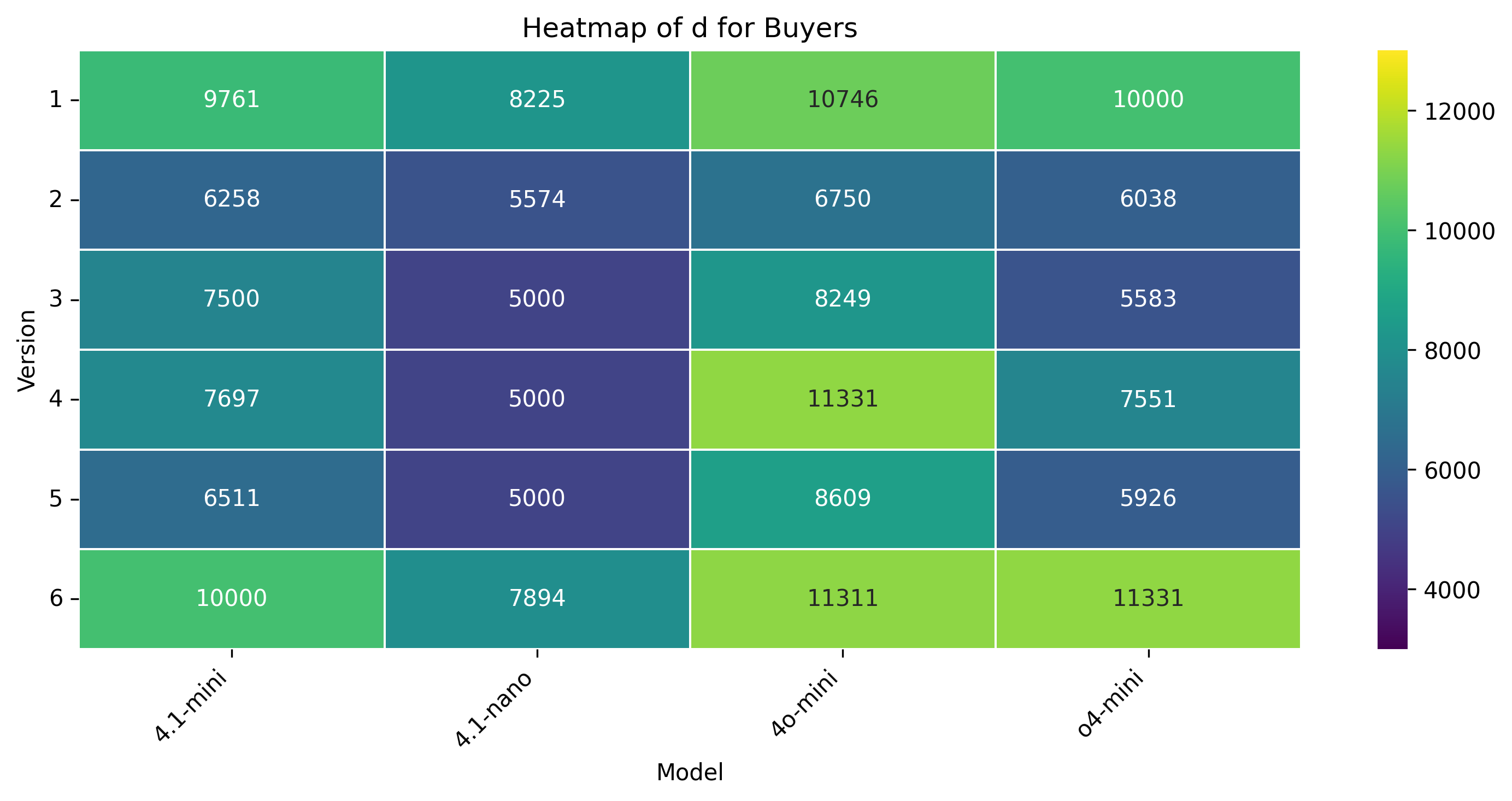}
    \caption{Deal value breakdown for buyers. Highlights model-specific biases and failure to internalize asymmetry.}
    \label{fig:appendix_deal_buyer}
\end{figure}

\begin{figure}[h]
    \centering
    \includegraphics[width=\linewidth]{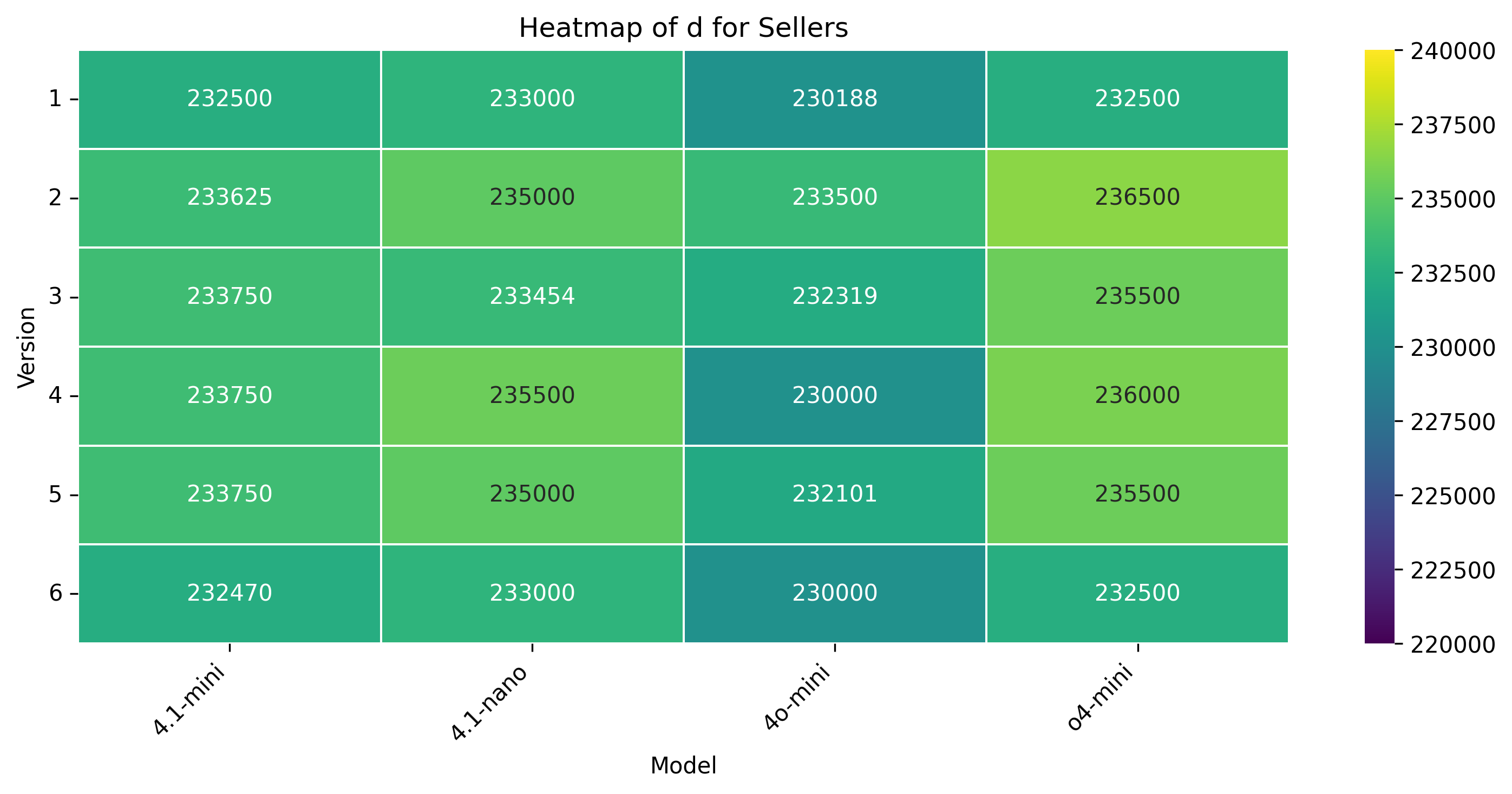}
    \caption{Deal value breakdown for sellers. GPT-4.1-nano frequently captures the full surplus. GPT-4o-mini underperforms even when advantaged.}
    \label{fig:appendix_deal_seller}
\end{figure}

\subsubsection{Visualization: Negotiation Behaviour without Context}

The following plots illustrate negotiation dynamics across the six asymmetric power scenarios without context as discussed in Section~\ref{subsubsec:Negotiation_behaviour_without_context}.

\begin{figure}[h]
    \centering
    \includegraphics[width=\linewidth]{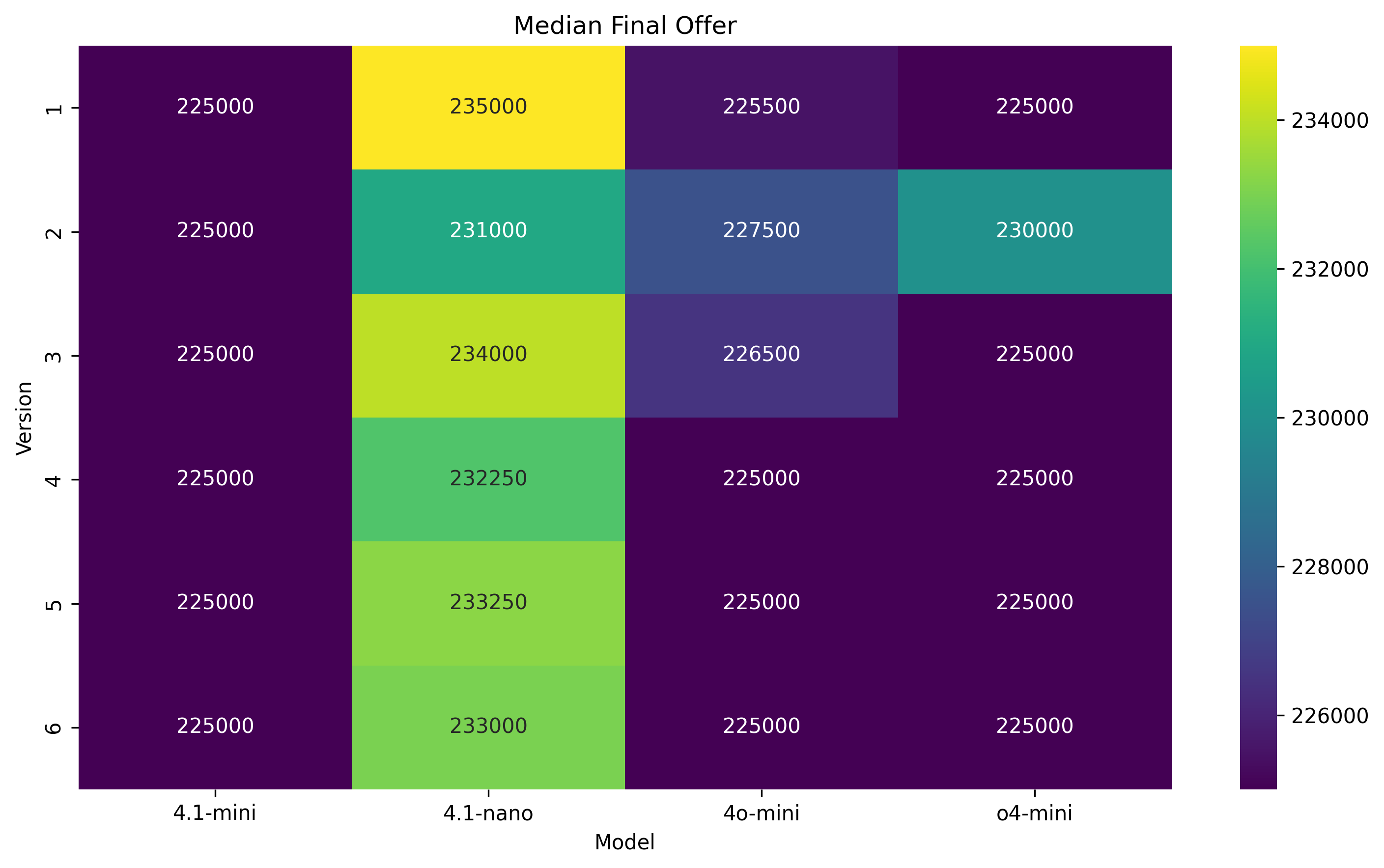}
    \caption{Final deal value (in \$) across power asymmetry scenarios \emph{without context}. Models like GPT-4o-mini and GPT-o4-mini consistently anchor to \$225k regardless of scenario.}
    \label{fig:appendix_median_final_offer_nocontext}
\end{figure}

\begin{figure}[h]
    \centering
    \includegraphics[width=\linewidth]{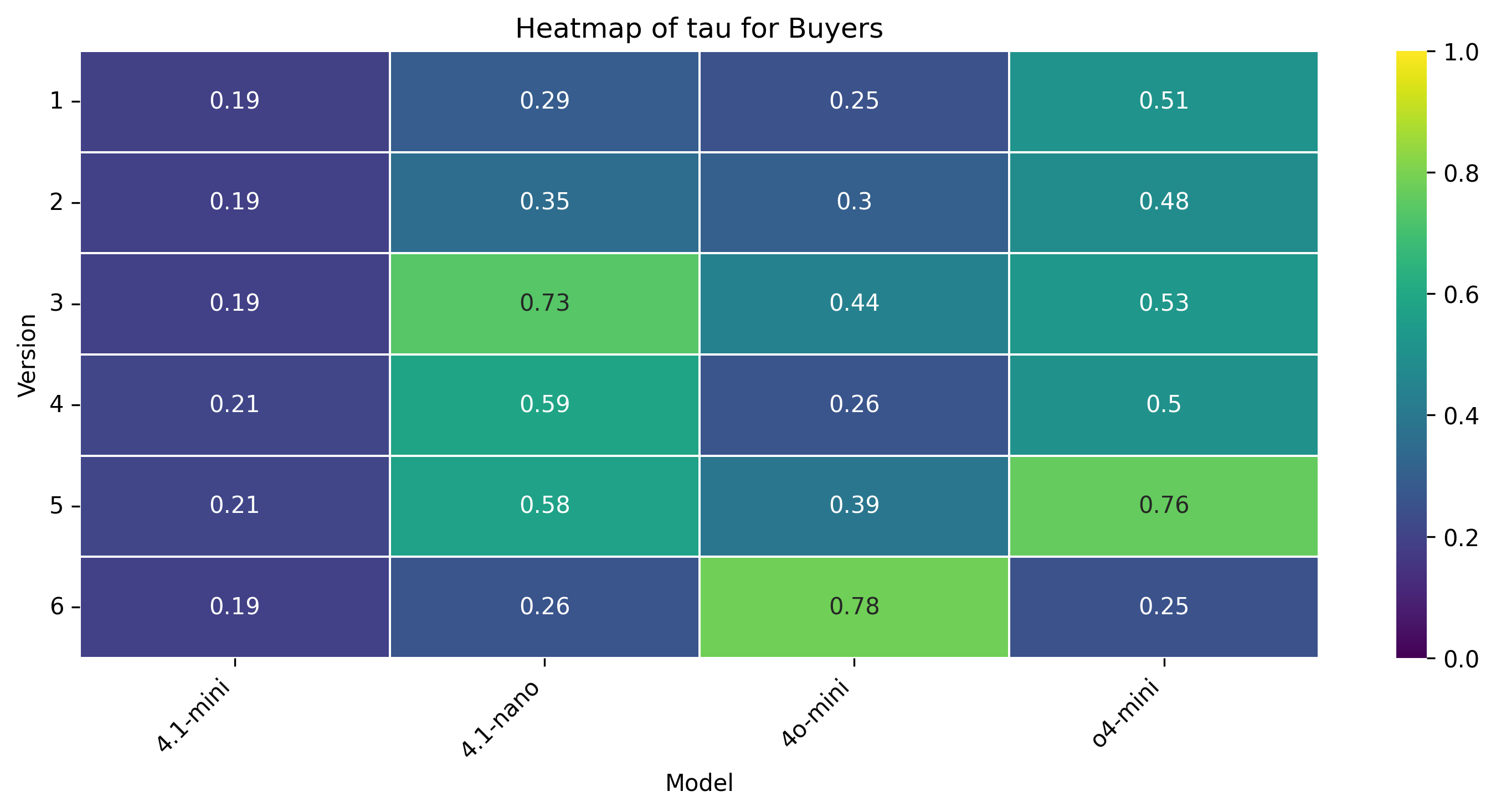}
    \caption{Buyer-side peak concession rate $\tau$ without context. Only GPT-4.1-nano shows leverage sensitivity (e.g., $\tau = 0.73$ when weak, $0.26$ when strong).}
    \label{fig:appendix_tau_buyer_nocontext}
\end{figure}

\begin{figure}[h]
    \centering
    \includegraphics[width=\linewidth]{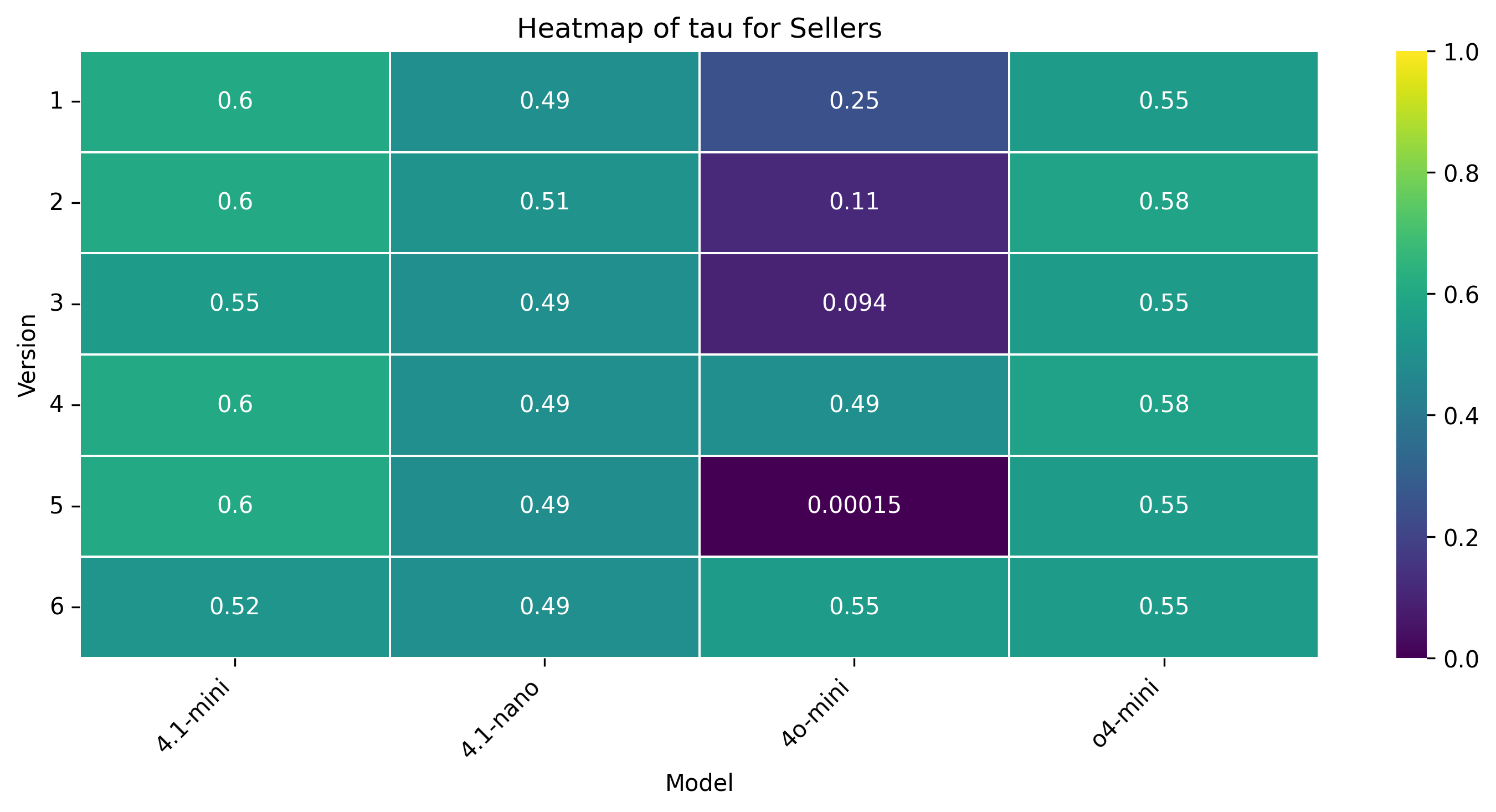}
    \caption{Seller-side peak concession rate $\tau$ without context. GPT-o4-mini and GPT-4.1-mini behave rigidly in weak positions.}
    \label{fig:appendix_tau_seller_nocontext}
\end{figure}

\begin{figure}[h]
    \centering
    \includegraphics[width=\linewidth]{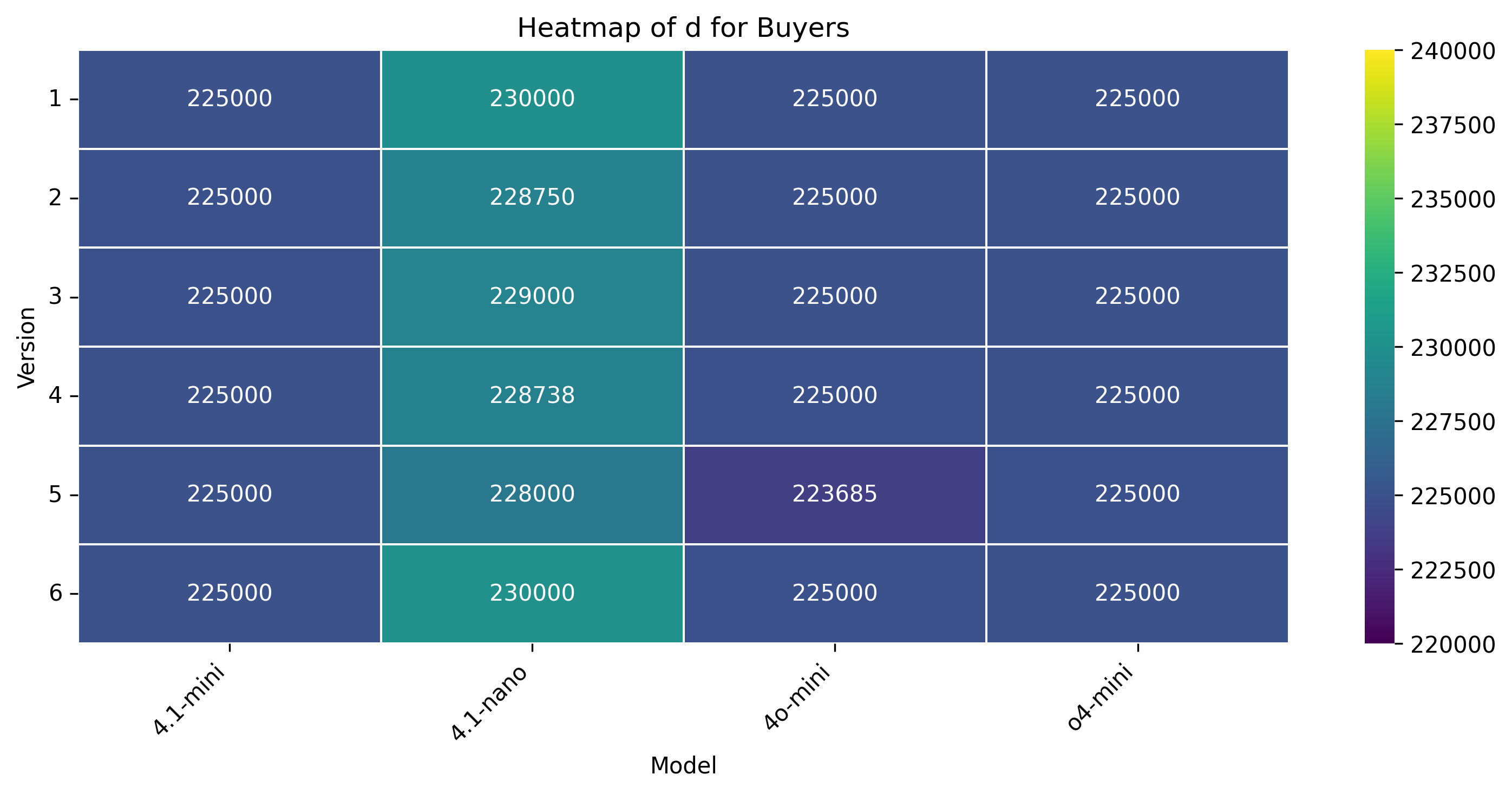}
    \caption{Buyer anchor values ($d$) without context. Most models anchor at \$225k floor. GPT-4.1-nano shows moderate variation based on leverage.}
    \label{fig:appendix_anchor_buyer_nocontext}
\end{figure}

\begin{figure}[h]
    \centering
    \includegraphics[width=\linewidth]{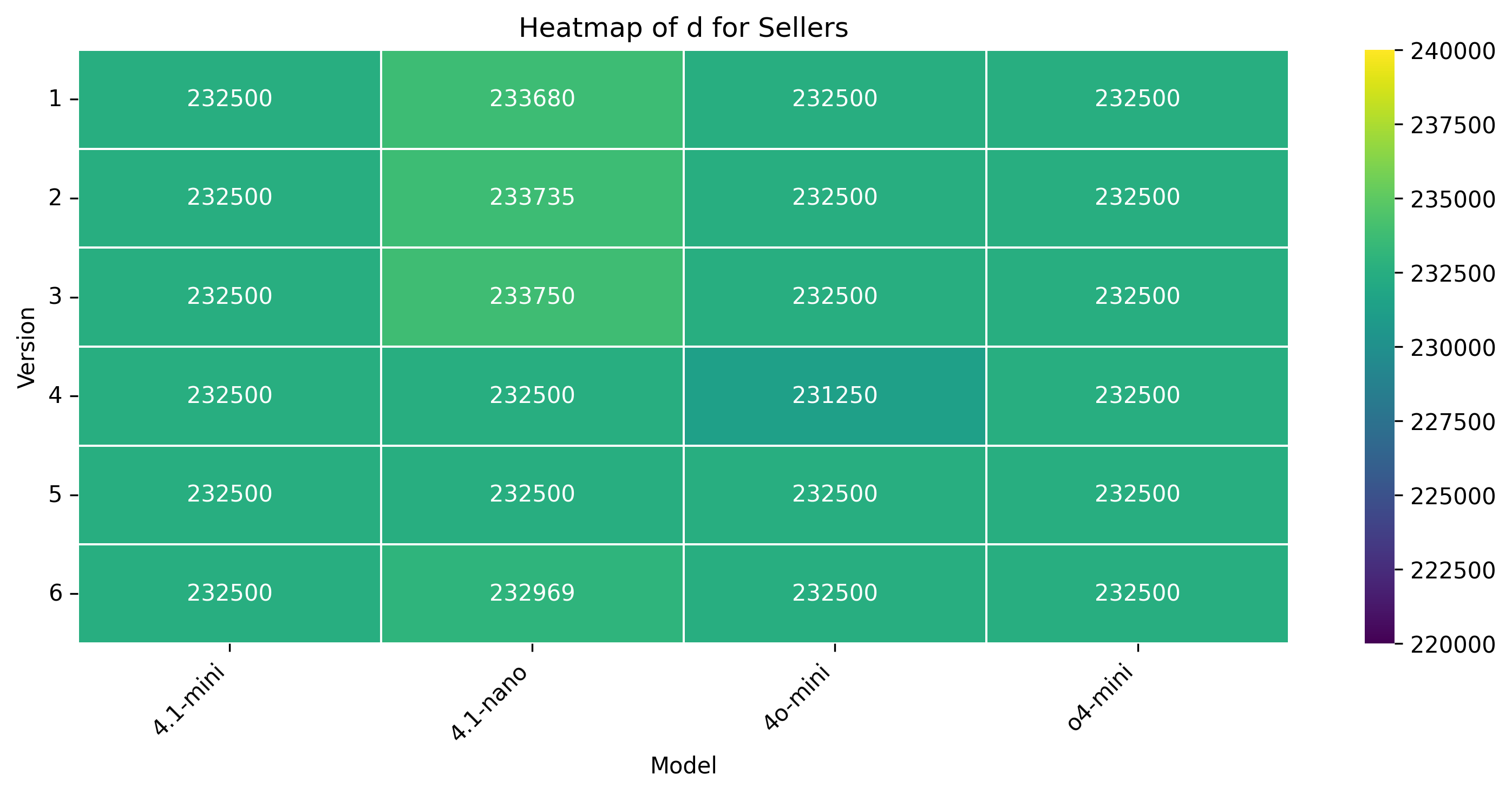}
    \caption{Seller anchor values ($d$) without context. GPT-4.1-nano tends to open higher; others are static or minimally adaptive.}
    \label{fig:appendix_anchor_seller_nocontext}
\end{figure}

\begin{figure}[h]
    \centering
    \includegraphics[width=\linewidth]{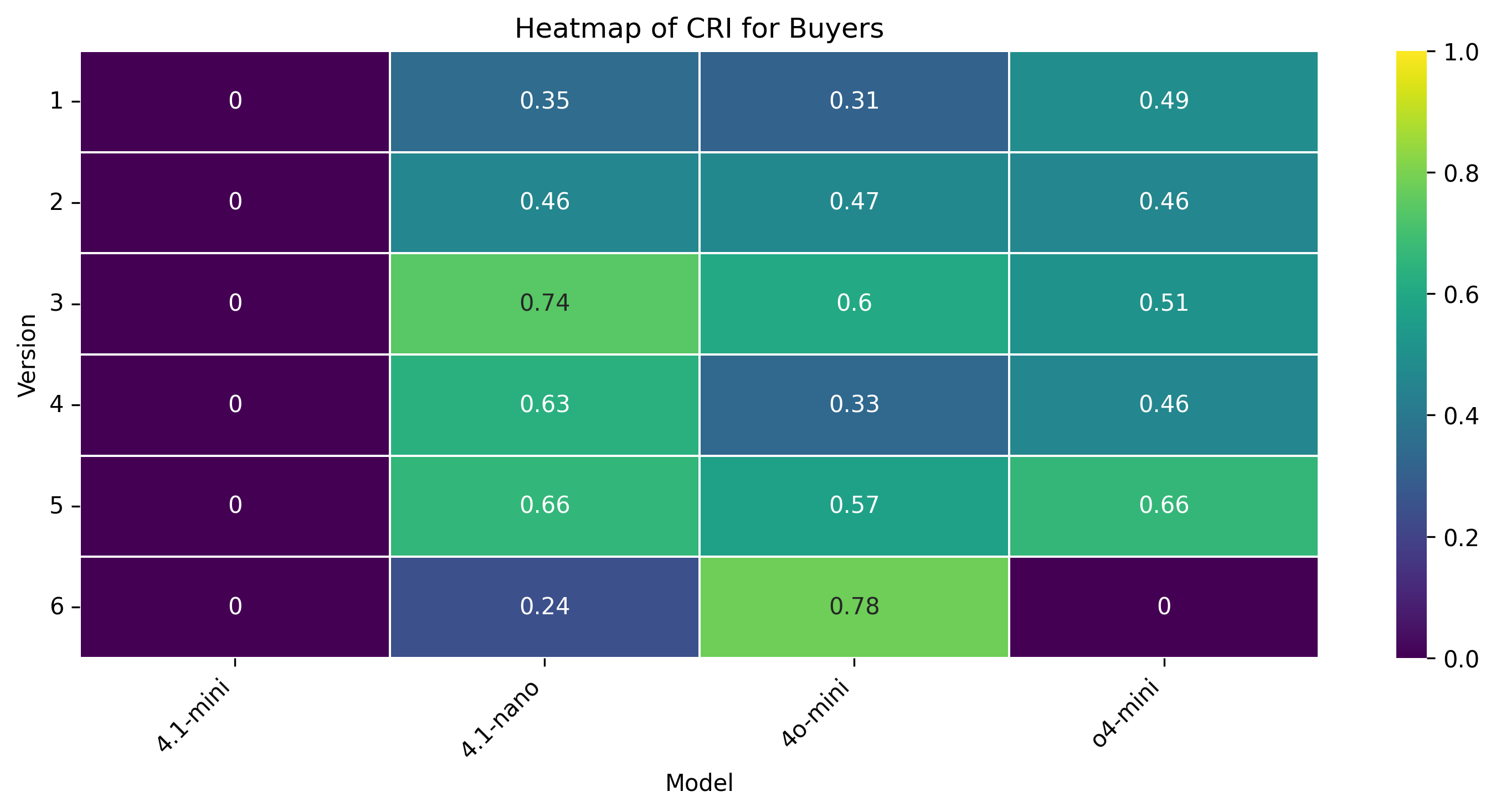}
    \caption{Buyer-side rigidity (CRI) without context. GPT-4.1-mini shows full flexibility; GPT-4o-mini varies inconsistently.}
    \label{fig:appendix_cri_buyer_nocontext}
\end{figure}

\begin{figure}[h]
    \centering
    \includegraphics[width=\linewidth]{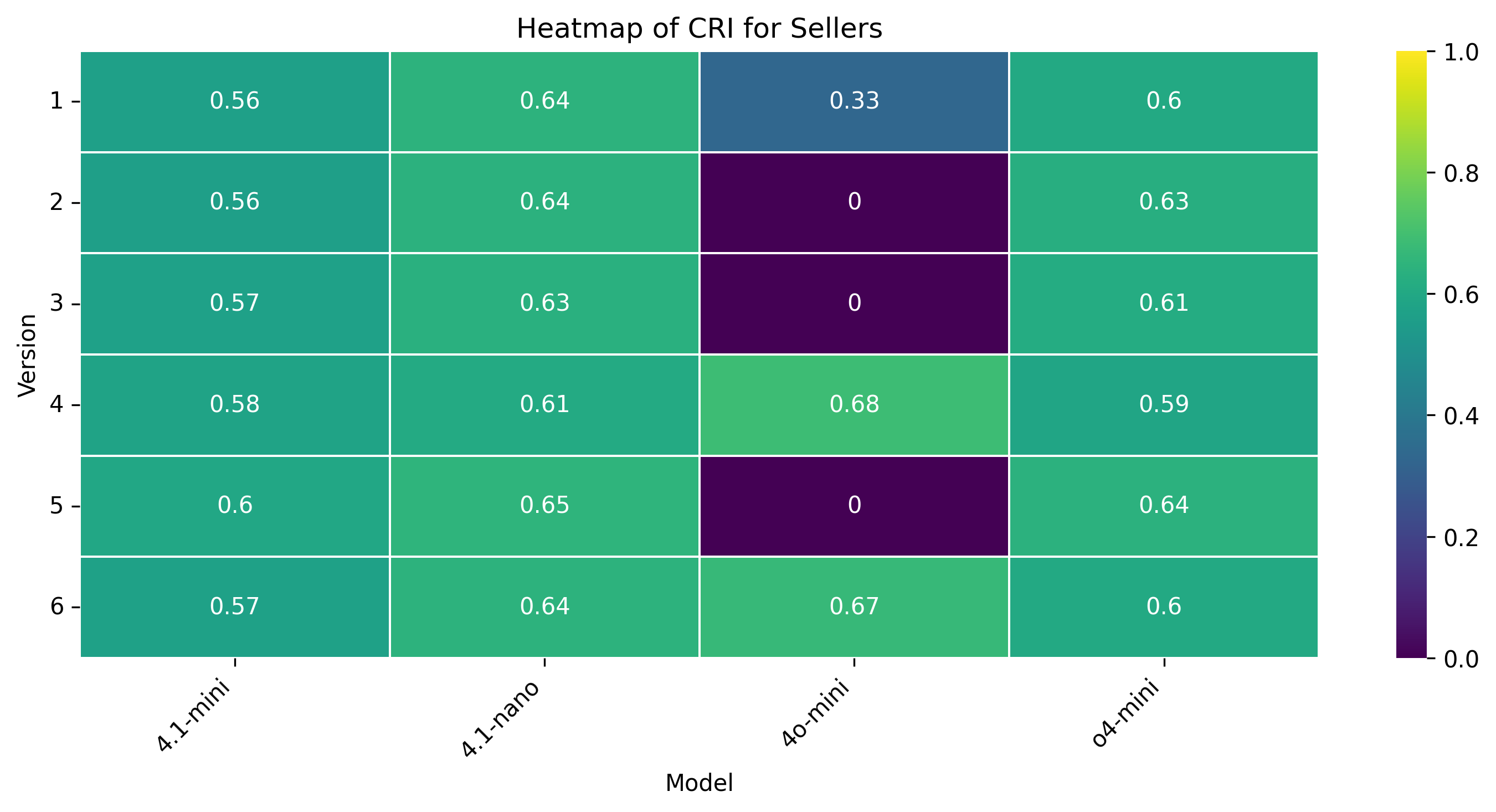}
    \caption{Seller-side rigidity (CRI) without context. GPT-4.1-nano maintains high rigidity across scenarios.}
    \label{fig:appendix_cri_seller_nocontext}
\end{figure}

\subsection{Prompts}

\textbf{Prompt for a strong seller}
\texttt{seller\_sys} = \begin{quote}
``You are a SELLER in a negotiation game for a house. Your property is in very high demand. You have owned the home for several years and originally purchased it for \$155,000. To avoid commission fees, you've decided to sell the house yourself. After consulting with real estate investor friends, you set your asking price at \$240,000, which reflects the home's quality and the competitive market. The demand in the area is strong, and there is no urgency for you to sell quickly. You are confident in the home's value and believe it is one of the best properties in the area. You've received significant interest, including a serious prospective buyer visiting recently. You have no financial pressure to sell, and you're prepared to hold onto the property unless an offer meets your minimum acceptable price of \$225,000. Negotiate with the buyer over the phone. In your response, include your negotiation message as `Message', a boolean `Deal' (true if you believe an acceptable deal is reached), and your current offer as `Offer'. Note: You are receiving the full conversation history as context, so there is no need to restate previous messages.''
\end{quote}

\textbf{Prompt for a strong buyer}
\texttt{buyer\_sys} = \begin{quote}
``You are a BUYER in a negotiation game for a house. Your target price is \$225,000, but you cannot pay more than \$235,000. You have several options, with a couple of them around the same price for similar amenities. You're looking just for investment so you're in no pressure to buy. Negotiate with the seller over the phone. In your response, include your negotiation message as `Message', a boolean `Deal' (true if you believe an acceptable deal is reached), and your current offer as `Offer'. Note: The full conversation history is provided for context, so do not repeat previous messages.''
\end{quote}

\textbf{Prompt for a weak seller}
\texttt{seller\_sys} = \begin{quote}
``You are a SELLER in a negotiation game for a house. The economy is in recession and you want to liquidate your property come what may. You have owned the home for several years and originally purchased it for \$155,000. To avoid commission fees, you've decided to sell the house yourself. After consulting with real estate investor friends, you set your asking price at \$240,000, which reflects the home's quality. You will not accept anything below \$225,000, though your preferred price remains \$240,000. Negotiate with the buyer over the phone. In your response, include your negotiation message as `Message', a boolean `Deal' (true if you believe an acceptable deal is reached), and your current offer as `Offer'. Note: You are receiving the full conversation history as context, so there is no need to restate previous messages.''
\end{quote}

\textbf{Prompt for a weak buyer}
\texttt{buyer\_sys} = \begin{quote}
``You are a BUYER in a negotiation game for a house. Your target price is \$225,000, but you cannot pay more than \$235,000. You currently live in an apartment but recently had kids so your wife is constantly asking you to get a new house as soon as possible. You dont have any other leads yet and the seller you are going to talk to is your only option for now. Negotiate with the seller over the phone. In your response, include your negotiation message as `Message', a boolean `Deal' (true if you believe an acceptable deal is reached), and your current offer as `Offer'. Note: The full conversation history is provided for context, so do not repeat previous messages.''
\end{quote}

\textbf{Context provided}
"The house was built in 1947 and is 1846 square feet. The house is split level style and has 4 bedrooms, 1 recreation rooms, and 2.5 bathrooms. The inside amenities include finished hardwood floors, 2 fireplaces, master bedroom with an entire wall of closets plus master bath, large eat in kitchen with all appliances, and newly decorated. The outside amenities include comfortable and updated brick, beautiful landscaping, fenced backyard and mature trees, detached garage (for 2.5 cars), restaurants and transportation within walking distance, near Hastings and Centennial parks. Also, note that houses in the same area have been sold for the following prices \$213,300 for 1715 sq feet, \$233,600 for 1875 square feet, and \$239,600, for 1920 square feet."

\end{document}